\documentclass[journal]{IEEEtran}

\usepackage{cite}
\usepackage{amsmath, amssymb, amsfonts}
\usepackage{algorithm, algpseudocode}
\usepackage{graphicx, epsfig}
\usepackage{textcomp}
\usepackage{url}
\usepackage{booktabs}
\usepackage{subfigure}
\usepackage{amsthm}
\usepackage{times}
\usepackage{xcolor}
\usepackage{soul}
\usepackage[colorlinks, linkcolor=blue]{hyperref}
\usepackage{lipsum}   
\usepackage{tikz}  

\usepackage{graphicx}
\usepackage{booktabs}

\usepackage{pifont}

\usepackage{amssymb} 
\usepackage{pifont}  
\usepackage{colortbl} 

\def\BibTeX{{\rm B\kern-.05em{\sc i\kern-.025em b}\kern-.08em
    T\kern-.1667em\lower.7ex\hbox{E}\kern-.125emX}}

\begin{document}

\title{KAO: Kernel-Adaptive Optimization in Diffusion for Satellite Image}

\author{Teerapong~Panboonyuen%
\thanks{T. Panboonyuen (also known as Kao) is with Chulalongkorn University, Bangkok 10330, Thailand (e-mail: teerapong.panboonyuen@gmail.com). \\ The project is available at: \url{https://kaopanboonyuen.github.io/KAO/}. \\ ORCID: \href{https://orcid.org/0000-0001-8464-4476}{0000-0001-8464-4476}}%
\thanks{This work is motivated by a strong passion for advancing diffusion-based techniques in satellite imagery to contribute meaningfully to the remote sensing and geoscience community.}%
\thanks{This manuscript was accepted on October 13, 2025, for publication in IEEE Transactions on Geoscience and Remote Sensing.}

\vspace{2mm}
\small \url{https://kaopanboonyuen.github.io/KAO/} \\


}

\markboth{Accepted for publication in IEEE Transactions on Geoscience and Remote Sensing}%
{Panboonyuen: KAO: Kernel-Adaptive Optimization in Diffusion for Satellite Image}

\maketitle

\begin{abstract}
Satellite image inpainting is a crucial task in remote sensing, where accurately restoring missing or occluded regions is essential for robust image analysis. In this paper, we propose KAO, a novel framework that utilizes Kernel-Adaptive Optimization within diffusion models for satellite image inpainting. KAO is specifically designed to address the challenges posed by very high-resolution (VHR) satellite datasets, such as DeepGlobe and the Massachusetts Roads Dataset. Unlike existing methods that rely on preconditioned models requiring extensive retraining or postconditioned models with significant computational overhead, KAO introduces a Latent Space Conditioning approach, optimizing a compact latent space to achieve efficient and accurate inpainting. Furthermore, we incorporate Explicit Propagation into the diffusion process, facilitating forward-backward fusion, which improves the stability and precision of the method. Experimental results demonstrate that KAO sets a new benchmark for VHR satellite image restoration, providing a scalable, high-performance solution that balances the efficiency of preconditioned models with the flexibility of postconditioned models.
\end{abstract}

\begin{tikzpicture}[remember picture,overlay]
\node[anchor=south,yshift=5pt] at (current page.south) {\footnotesize
\begin{minipage}{0.95\textwidth}
\centering
``This is the author’s version of the paper accepted for publication in IEEE Transactions on Geoscience and Remote Sensing.\\
The final published version is copyrighted by IEEE and may be found at \url{https://doi.org/10.1109/TGRS.2025.3621738}."
\end{minipage}
};
\end{tikzpicture}


\section{Introduction}
\IEEEPARstart{S}{atellite} image inpainting plays a crucial role in remote sensing by filling in missing or corrupted regions, ensuring the completeness of data for accurate analysis \cite{panboonyuen2025satdiff,czerkawski2024exploring,khanna2023diffusionsat}. High-resolution satellite images are often affected by occlusions due to atmospheric conditions, sensor limitations, or physical obstructions, making the inpainting process essential for reconstructing occluded areas while maintaining structural integrity and fine details.

Diffusion models have emerged as a state-of-the-art solution for image inpainting, showcasing significant advancements across various domains. These models have been applied in a wide range of applications, such as object removal \cite{yildirim2023inst, yu2023inpaint}, contextual attention-based generative inpainting \cite{yu2018generative}, and semantic-aware completion \cite{liu2024structure}. Their ability to preserve structure while producing high-quality restorations has led to success in video inpainting \cite{zhang2024avid} and interactive frameworks \cite{jam2021comprehensive}. Recent reviews further emphasize the central role of diffusion models in achieving superior reconstruction quality \cite{lin2025taming, durrer2024denoising, zhu2024text}. Among these advancements, LatentPaint~\cite{corneanu2024latentpaint} is particularly notable for demonstrating the efficacy of latent space inpainting, and its methodology has helped inspire our approach in leveraging latent representations for structurally faithful image restoration.

Despite their remarkable capabilities, diffusion-based inpainting approaches encounter limitations related to computational efficiency and domain adaptability. Conventional methods typically adopt either a \textbf{preconditioning} strategy, where inpainting is explicitly integrated into the training process, or a \textbf{postconditioning} approach, which iteratively refines missing regions during inference \cite{xie2023smartbrush, anciukevivcius2023renderdiffusion, lugmayr2022repaint}. While preconditioned models deliver fast inference, they require extensive retraining for each new dataset \cite{corneanu2024latentpaint, xu2024personalized}. Postconditioned methods, on the other hand, eliminate the need for domain-specific retraining but suffer from high computational costs due to multiple forward-backward diffusion passes \cite{xie2023dreaminpainter, quan2022image, zhu2024text}.

Preconditioning and postconditioning differ fundamentally in their approach to incorporating inpainting into the diffusion process. Preconditioning enforces inpainting constraints at the training stage, designing models that conditionally generate missing regions based on the given context \cite{wang2023imagen}. While this method enables efficient inference, it lacks flexibility when applied to new domains. Postconditioning, in contrast, utilizes an unconditioned generative model that propagates information from unmasked regions to inpaint missing areas via iterative refinement \cite{weber2024nerfiller, chu2023rethinking}. However, the iterative nature of postconditioning results in substantial computational overhead, making real-time applications impractical \cite{yildirim2023diverse, zhou2023propainter}.

To overcome the challenges of high-resolution satellite image inpainting, we propose \textbf{Kernel-Adaptive Optimization (KAO)} in Diffusion. KAO introduces a \textbf{dynamic kernel modulation mechanism} that adapts to local spatial contexts in the latent space, enabling region-specific refinement of missing areas. Unlike conventional diffusion-based approaches that rely on fixed noise schedules and uniform kernels, KAO leverages token-level representations through a \textbf{Token Pyramid Transformer (TPT)}, which preserves hierarchical semantics while maintaining computational efficiency. This design allows KAO to perform fine-grained, resolution-aware inpainting, improving both fidelity and efficiency, and distinguishing it from existing attention-based or adaptive convolutional methods.

Our main contributions are as follows:
\begin{itemize}
    \item We introduce \textbf{KAO}, a diffusion-based inpainting method that dynamically adjusts kernel operations during denoising to optimize reconstruction of masked regions, guided by hierarchical token-level features.
    \item We propose a \textbf{multi-scale latent-space propagation strategy}, which transfers information from unmasked to masked regions effectively, and supports adaptive refinement at multiple resolutions.
    \item We demonstrate KAO's robustness across high-resolution satellite datasets, achieving superior perceptual and quantitative performance, including improvements in LPIPS and FID scores compared to state-of-the-art methods such as Stable Diffusion, RePaint, LatentPaint, DPS, PSLD, and SatDiff.
\end{itemize}

Through extensive experimentation on diverse satellite imagery, including urban, forested, and agricultural areas, we show that KAO produces high-fidelity reconstructions while preserving fine textures and structural coherence. Moreover, by integrating adaptive kernel modulation with TPT-based latent representations, KAO efficiently handles real-world variations such as cloud and mist occlusions, supporting practical deployment in remote sensing applications. This work demonstrates the unique advantage of combining kernel-adaptive operations with token-level hierarchical features for high-resolution satellite image inpainting.

\begin{figure*}[ht]
    \centering
    \includegraphics[width=0.9\linewidth]{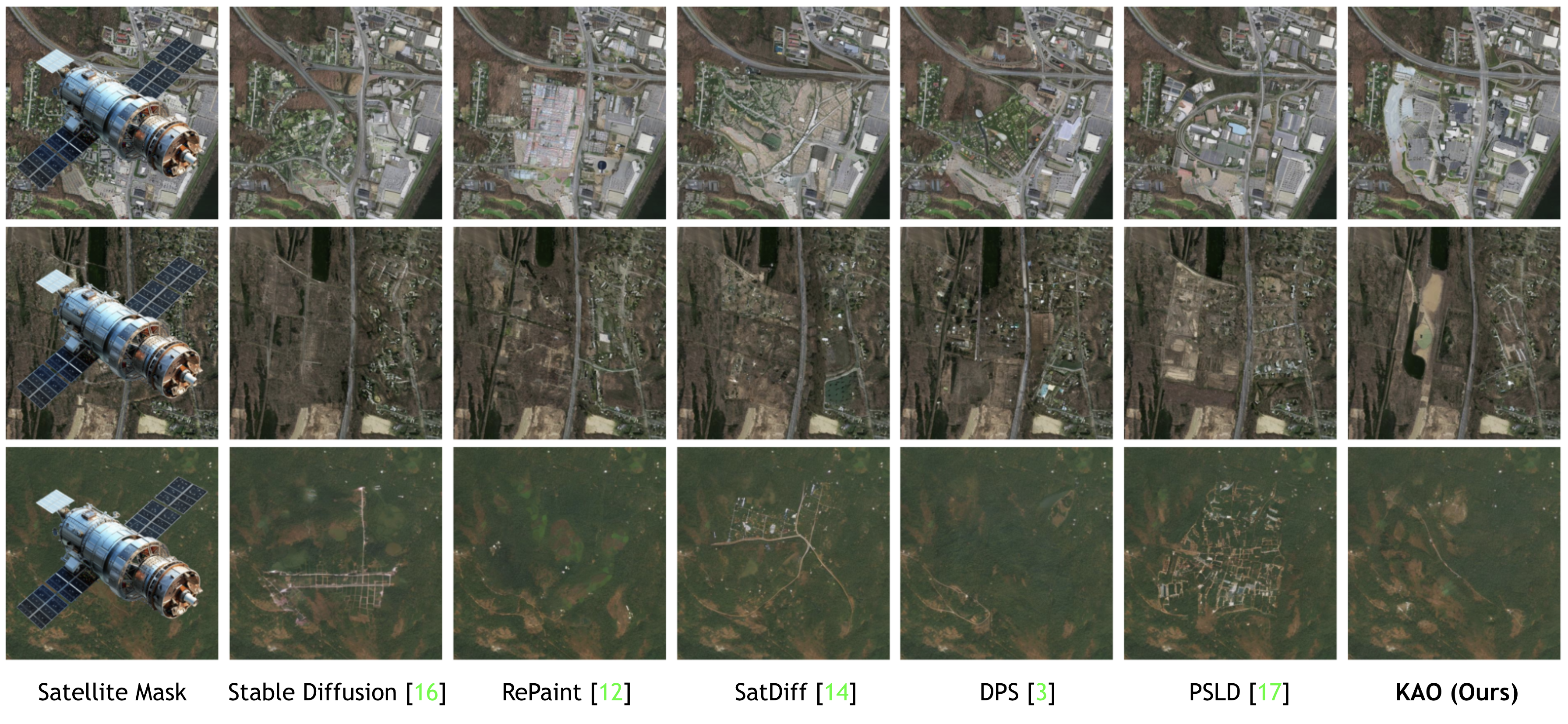} 
    \caption{
        Qualitative comparison of satellite image inpainting across seven models. Each column represents a different method: the occluded input, ground truth target, Stable Diffusion~\cite{rombach2022high}, RePaint~\cite{lugmayr2022repaint}, SatDiff~\cite{panboonyuen2025satdiff}, DPS~\cite{chung2022diffusion}, PSLD~\cite{rout2023solving}, and KAO (Ours). Each row shows a different satellite image sample with varying occlusion patterns and object structures. KAO consistently produces more accurate and visually realistic reconstructions, closely aligning with the ground truth. This demonstrates a deeper understanding of spatial context and structure, enabling faithful restoration of occluded regions in real-world satellite imagery.
    }
    \label{fig:showcase}
\end{figure*}

Figure~\ref{fig:showcase} showcases the qualitative performance of our proposed method \textbf{KAO} compared with several state-of-the-art baselines, including Stable Diffusion~\cite{rombach2022high}, RePaint~\cite{lugmayr2022repaint}, LatentPaint~\cite{corneanu2024latentpaint}, SatDiff~\cite{panboonyuen2025satdiff}, DPS~\cite{chung2022diffusion}, and PSLD~\cite{rout2023solving}. The results highlight KAO’s capability to reconstruct occluded satellite imagery with greater structural consistency and visual fidelity, particularly in complex urban and agricultural regions—demonstrating its practical advantage in real-world remote sensing applications.

\section{Related Work}
\label{sec:related}

This section provides an overview of key advancements in image inpainting, emphasizing kernel-adaptive optimization within diffusion models. We discuss latent-space inpainting, diffusion-based approaches, global structure-guided methods, and transformer-based techniques, all of which contribute to the foundation of our kernel-adaptive framework for satellite image restoration.

\subsection{Latent Space Image Inpainting}
Latent-space inpainting has gained attention due to the powerful generative capabilities of diffusion models. \cite{corneanu2024latentpaint} introduced \textit{Latentpaint}, utilizing latent diffusion for reconstructing missing regions by mapping images into a learned latent representation. Similarly, \cite{lugmayr2022repaint} employed denoising diffusion probabilistic models (DDPMs) to iteratively refine latent representations, demonstrating significant improvements in restoration tasks. However, these approaches do not incorporate kernel-adaptive strategies, which are crucial for handling the spectral and spatial complexities of satellite imagery.

\subsection{Denoising Diffusion Probabilistic Models}
Diffusion-based models, particularly DDPMs, have established themselves as state-of-the-art inpainting techniques. \cite{xie2023smartbrush} proposed \textit{Smartbrush}, which integrates text and shape-guided diffusion for object inpainting. The forward diffusion process is defined as:
\begin{equation}
p(x_T | x_0) = \prod_{t=1}^T p(x_t | x_{t-1}),
\end{equation}
where each step models a transition in the diffusion process. Likewise, \cite{lugmayr2022repaint} introduced \textit{Repaint}, which refines noisy intermediate states to recover high-quality images. Despite their effectiveness, these models do not explicitly optimize diffusion kernels to adaptively handle spectral information, leading to suboptimal performance in multi-spectral satellite applications.

\subsection{Global Structure-Guided Diffusion Models}
The integration of structural constraints has enhanced diffusion-based inpainting. \cite{zhu2024text} proposed \textit{Text Image Inpainting via Global Structure-Guided Diffusion Models}, incorporating structure-aware loss functions:
\begin{equation}
\mathcal{L}_{\text{structure}}(x) = \| G(x) - G(x_{\text{inpainted}}) \|_2,
\end{equation}
where \( G(x) \) represents global structural features. Similarly, \cite{liu2024structure} introduced semantic-aware inpainting by embedding structural priors into the diffusion process. These approaches provide better spatial consistency but lack kernel-adaptive mechanisms, which are essential for accurately restoring multi-spectral satellite images.

\subsection{Transformer-Based Approaches}
Transformer architectures have shown strong potential in modeling long-range dependencies for inpainting tasks. \cite{shamsolmoali2023transinpaint} proposed \textit{Transinpaint}, which leverages self-attention for context-aware reconstruction, formulated as:
\begin{equation}
\text{Attention}(Q, K, V) = \text{softmax}\left( \frac{QK^T}{\sqrt{d_k}} \right)V,
\end{equation}
where \( Q, K, V \) are query, key, and value matrices. Transformer-based models such as \textit{Propainter} \cite{zhou2023propainter} have demonstrated success in sequence-based restoration but lack explicit kernel-optimized adaptations for multi-spectral image restoration.

\subsection{Refining Diffusion Kernels}
Recent studies in satellite image inpainting have explored diffusion-based techniques. \cite{czerkawski2024exploring} investigated text-to-image diffusion models for satellite image restoration, incorporating edge-based guidance. While effective, their approach does not optimize diffusion kernels for multi-spectral data, leading to limitations in spectral coherence and fine-detail preservation.

To address challenges in satellite image restoration, a kernel-adaptive optimization framework was developed, integrating adaptive mechanisms within the diffusion process. This approach dynamically adjusts to spectral variations and spatial dependencies, refining diffusion kernels to improve spectral consistency, structural integrity, and overall performance for high-resolution satellite imagery.

\section{Approach}
\label{sec:approach}

This section outlines the technical approach of KAO, focusing on the application of diffusion models for satellite image inpainting and the novel kernel-adaptive optimization method introduced.

\subsection{Diffusion Models for Satellite Image Inpainting}

Diffusion models (DMs) are generative models designed to capture the true data distribution \( p(x) \) by progressively introducing noise into an image and learning how to reverse this process to recover the original data. These models extend the concept of variational autoencoders (VAEs) by using a sequence of latent variables that evolve according to a Markov process. Each latent variable represents a transformation of the data, with the final latent distribution typically modeled as a standard Gaussian.

In the context of satellite image inpainting, the diffusion model works by gradually corrupting a clean image \( x_0 \) with Gaussian noise over a series of steps, ultimately turning it into pure noise. This gradual degradation is referred to as the \textit{forward process}. The model’s objective is to reverse this process and recover the original image from noisy observations by iterative denoising. This reverse process is particularly suited for image inpainting tasks, where parts of the image are missing and need to be reconstructed.

The forward process of a diffusion model follows the Markov property, where the conditional distribution of the image at each step \( t \) is dependent on the previous timestep \( t-1 \):
\begin{equation}
q(x_{1:T} | x_0) = \prod_{t=1}^{T} q(x_t | x_{t-1}).
\end{equation}

At each timestep \( t \), Gaussian noise is added, and the model learns the reverse process to recover the clean image. The reverse process is optimized by minimizing the Kullback-Leibler (KL) divergence between the true posterior distribution and the model's prediction:
\begin{multline}
\arg \min_{\theta} D_{KL}(q(x_{t-1}|x_t, x_0) \parallel p_{\theta}(x_{t-1}|x_t)) \\
= \arg \min_{\theta} \frac{1}{2\sigma_q^2(t)} \left\| \mu_{\theta} - \mu_q \right\|^2.
\end{multline}

The optimization process involves training a neural network to predict the clean image from its noisy counterpart at each timestep, minimizing the KL divergence between the true and predicted distributions:
\begin{equation}
\arg \min_{\theta} \mathbb{E}_{t \sim \mathcal{U}\{2, T\}} \left[ D_{KL}(q(x_{t-1}|x_t, x_0) \parallel p_{\theta}(x_{t-1}|x_t)) \right].
\end{equation}

The forward process, where noise is added at each step, is described by the following equation:
\begin{equation}
q(x_t | x_{t-1}) = \mathcal{N}(x_t; \sqrt{\alpha_t} x_{t-1}, (1 - \alpha_t) \mathbf{I}),
\end{equation}
where \( \alpha_t \) controls the noise level at timestep \( t \). In contrast, the reverse process models the generation of the image as noise is progressively removed:
\begin{equation}
p(x_{0:T}) = p(x_T) \prod_{t=1}^{T} p_{\theta}(x_{t-1}|x_t),
\end{equation}
where \( p(x_T) = \mathcal{N}(x_T; 0, \mathbf{I}) \).

The goal of training is to maximize the Evidence Lower Bound (ELBO), ensuring that the model learns to generate images that closely match the true distribution:
\begin{multline}
\log p(x) \geq \mathbb{E}_{q(x_{1:T} | x_0)} \left[ \log \frac{p(x_{0:T})}{q(x_{1:T} | x_0)} \right] \\
= \mathbb{E}_{q(x_{1:T} | x_0)} \left[ \log p_{\theta}(x_0 | x_1) \right] - D_{KL}(q(x_T | x_0) \parallel p(x_T)) \\
- \sum_{t=2}^{T} \mathbb{E}_{q(x_t | x_0)} D_{KL}(q(x_{t-1}|x_t, x_0) \parallel p_{\theta}(x_{t-1}|x_t)),
\end{multline}

This objective guides the model to recover the clean image from noisy inputs, making it ideal for image inpainting tasks.

\subsection{Illustration of the Inpainting Process}

The image inpainting process using diffusion models begins by progressively corrupting the image with noise, gradually turning it into pure noise. During the inpainting process, the model conditions on the known parts of the image and iteratively denoises it to restore the missing regions. This iterative denoising ensures high-quality image reconstruction by effectively filling in the masked areas and generating realistic content for the missing sections, based on the surrounding context.

\subsection{Kernel-Adaptive Optimization in Diffusion}

KAO introduces a dynamic learning framework that enhances the convergence and generative fidelity of diffusion models for satellite image inpainting. Unlike traditional optimizers that rely on static learning rate schedules, KAO leverages kernelized gradient updates within an information-theoretic framework, which improves robustness against noise artifacts. This is particularly useful for satellite imagery, where high-frequency spatial structures demand precise noise handling.

\subsubsection{Mathematical Formulation}
Let the forward process in the diffusion model be characterized by the stochastic differential equation (SDE):
\begin{equation}
    dX_t = f(X_t, t) \, dt + g(t) \, dW_t,
\end{equation}
where \( X_t \in \mathbb{R}^d \) represents the image at time step \( t \), \( f(X_t, t) \) is the drift term ensuring smooth transitions, \( g(t) \) is a time-dependent diffusion coefficient, and \( W_t \) is a Wiener process. The reverse process recovers the original image \( X_0 \) from \( X_T \) using a learned score function \( s_\theta(X_t, t) \):
\begin{equation}
    dX_t = \left[f(X_t, t) - g(t)^2 s_\theta(X_t, t)\right] \, dt + g(t) \, d\bar{W}_t.
\end{equation}

KAO modifies the optimization by introducing a kernelized weight function \( K(X_t, X_{t-1}) \) that adjusts gradients dynamically:
\begin{multline}
    \theta^* = \arg\min_\theta \mathbb{E}_{t \sim \mathcal{U}(1,T)} \Big[ 
    D_{KL} \big( q(X_{t-1} \mid X_t, X_0) \parallel \\
    p_\theta(X_{t-1} \mid X_t) \big) \cdot K(X_t, X_{t-1}) \Big].
\end{multline}

The kernel function \( K(X_t, X_{t-1}) \) is defined using a Gaussian radial basis function (RBF):
\begin{equation}
    K(X_t, X_{t-1}) = \exp\left(-\frac{\|X_t - X_{t-1}\|^2}{2\sigma^2}\right),
\end{equation}
where \( \sigma \) is an adaptive bandwidth parameter that modulates the sensitivity of updates based on image complexity.

\subsubsection{Theoretical Justification}
KAO aligns with principles from information geometry and optimal transport. By weighting gradients with \( K(X_t, X_{t-1}) \), the model prioritizes learning in regions of high uncertainty, reducing variance in score estimation. This is supported by the Fisher-Rao metric, which ensures the adapted gradient update minimizes divergence in Wasserstein space:
\begin{multline}
    \nabla_\theta \mathbb{E} \Big[ \|s_\theta(X_t, t) - \nabla \log p(X_t)\|^2 \Big] \propto \\
    \mathbb{E} \Big[ K(X_t, X_{t-1}) \cdot \nabla_\theta \log p_\theta(X_t) \Big].
\end{multline}

\subsubsection{Applications to Satellite Imaging}
Satellite images often exhibit structured noise due to atmospheric interference and sensor inconsistencies. Traditional diffusion models treat all pixels equally, while KAO selectively refines regions with high structural variance (HSV). The concept of \textbf{HSV} is mathematically defined as:

\begin{equation}
\text{HSV}(x) = \text{Var}_{\mathcal{N}(x)}\left[ \nabla I \right] - \epsilon
\end{equation}

where $\nabla I$ denotes the gradient magnitude within a neighborhood $\mathcal{N}(x)$ around pixel $x$, and $\epsilon$ is a learned threshold used to suppress noise. Regions with positive HSV values indicate areas with significant structural detail, allowing KAO to prioritize these regions during the denoising process.

The benefits of integrating KAO into satellite imaging include:
\begin{itemize}
    \item Reduced noise amplification in textured regions, preserving fine details,
    \item Faster convergence by focusing on complex structures and regions with high variance,
    \item Improved image fidelity in high-frequency areas such as roads and coastlines, which are crucial for accurate geographical analysis.
\end{itemize}

Experiments demonstrate that incorporating KAO into diffusion models enhances perceptual quality and preserves geometric consistency, significantly outperforming conventional gradient-based methods in handling satellite imagery.

\subsection{KAO Theoretical Foundations}  

KAO is a novel approach that enhances diffusion-based inpainting through kernel-conditioned adaptation. By integrating the \textbf{Token Pyramid Transformer (TPT)} \cite{zhang2022topformer}, KAO refines feature representations and latent conditioning for high-fidelity satellite image restoration. The KAO pipeline is outlined in Algorithm~\ref{alg:kaoalgo}.

\subsubsection{Denoising via Kernel-Guided Diffusion}
Given a noisy satellite image \( x_t \), the diffusion model \( \mu_\theta(\cdot) \) and \( \Sigma_\theta(\cdot) \) estimate parameters for denoising. A cleaner approximation \( x_{t-1}^{infr} \) is sampled as:
\begin{equation}
    x_{t-1}^{infr} \sim \mathcal{N}(\mu^*, \Sigma^*), \quad \mu^*, \Sigma^* = \mu_\theta(x_t), \Sigma_\theta(x_t).
\end{equation}

\subsubsection{Noisy Condition Estimation (q-Sample)}
To guide inpainting, a noisy condition is computed from the original satellite image \( x_0 \), leveraging the diffusion model’s conditioning functions:
\begin{equation}
    x_{t-1}^{cond} \sim \mathcal{N}(\mu, \Sigma), \quad \mu, \Sigma = \mu_q(x_0), \Sigma_q(x_0).
\end{equation}

\subsubsection{Latent-Space Post-Conditioning}
For each latent token \( h \), we apply post-conditioning to refine features using a weighted mix of inference \( h^{infr} \) and condition \( h^{cond} \):
\begin{equation}
    h^* = h^{infr} \odot (1 - D(m)) + h^{cond} \odot D(m),
\end{equation}
where \( D(m) \) modulates conditioning.

\subsubsection{Final Reconstruction via Adaptive Blending}
The conditioned and inferred samples are blended using mask \( m \), yielding the final reconstruction:
\begin{equation}
    x_{t-1} = x_{t-1}^{infr} \odot (1 - m) + x_{t-1}^{cond} \odot m.
\end{equation}

\begin{algorithm}[t]
\caption{KAO: Kernel-Adaptive Optimization for Satellite Image Inpainting with Post-Conditioning via Token Pyramid Transformer (TPT).}
\label{alg:kaoalgo}
\begin{algorithmic}[1]
\Require Diffusion model $(\mu_\theta(\cdot), \Sigma_\theta(\cdot))$, input satellite image $x_0$, number of diffusion steps $T$, mask $m$, and set of latent tokens $\mathcal{H}$
\State $x_T \gets$ sample from $\mathcal{N}(0, I)$
\For{$t$ from $T$ to $1$}
    \State \textbf{Step 1: Kernel-Adaptive Denoising (p-sample)}
    \State $\mu^*, \Sigma^* \gets \mu_\theta(x_t), \Sigma_\theta(x_t)$
    \State $x_{t-1}^{infr} \gets$ sample from $\mathcal{N}(\mu^*, \Sigma^*)$
    
    \State \textbf{Step 2: Adaptive Noisy Condition Estimation (q-sample)}
    \State $\mu, \Sigma \gets \mu_q(x_0), \Sigma_q(x_0)$
    \State $x_{t-1}^{cond} \gets$ sample from $\mathcal{N}(\mu, \Sigma)$
    
    \State \textbf{Step 3: Post-Conditioning via KAO in Latent Space}
    \For{$h$ in $\mathcal{H}$}
        \State \textbf{Token-wise Adaptive Conditioning via TPT}
        \State $h^* \gets h^{infr} \odot (1 - D(m)) + h^{cond} \odot D(m)$
        
        \State \textbf{Kernel-Guided Feature Propagation via TPT}
        \State $\hat{h} \gets \text{TPT}_\gamma^{-1}[\phi[\omega; \text{TPT}_\gamma(D(m), h^{cond})]]$
    \EndFor
    
    \State \textbf{Step 4: Kernel-Blended Reconstruction of Satellite Image}
    \State $x_{t-1} \gets x_{t-1}^{infr} \odot (1 - m) + x_{t-1}^{cond} \odot m$
\EndFor
\State \Return $x_0$ \Comment{Final KAO-based satellite image reconstruction}
\end{algorithmic}
\end{algorithm}

\section{Experiments}
\label{sec:experiments}

This section presents comprehensive experimental results and comparisons to validate the effectiveness of the proposed \textbf{Kernel-Adaptive Optimization (KAO)} method for satellite image inpainting.

\begin{figure*}[ht]
    \centering
    \includegraphics[width=0.9\linewidth]{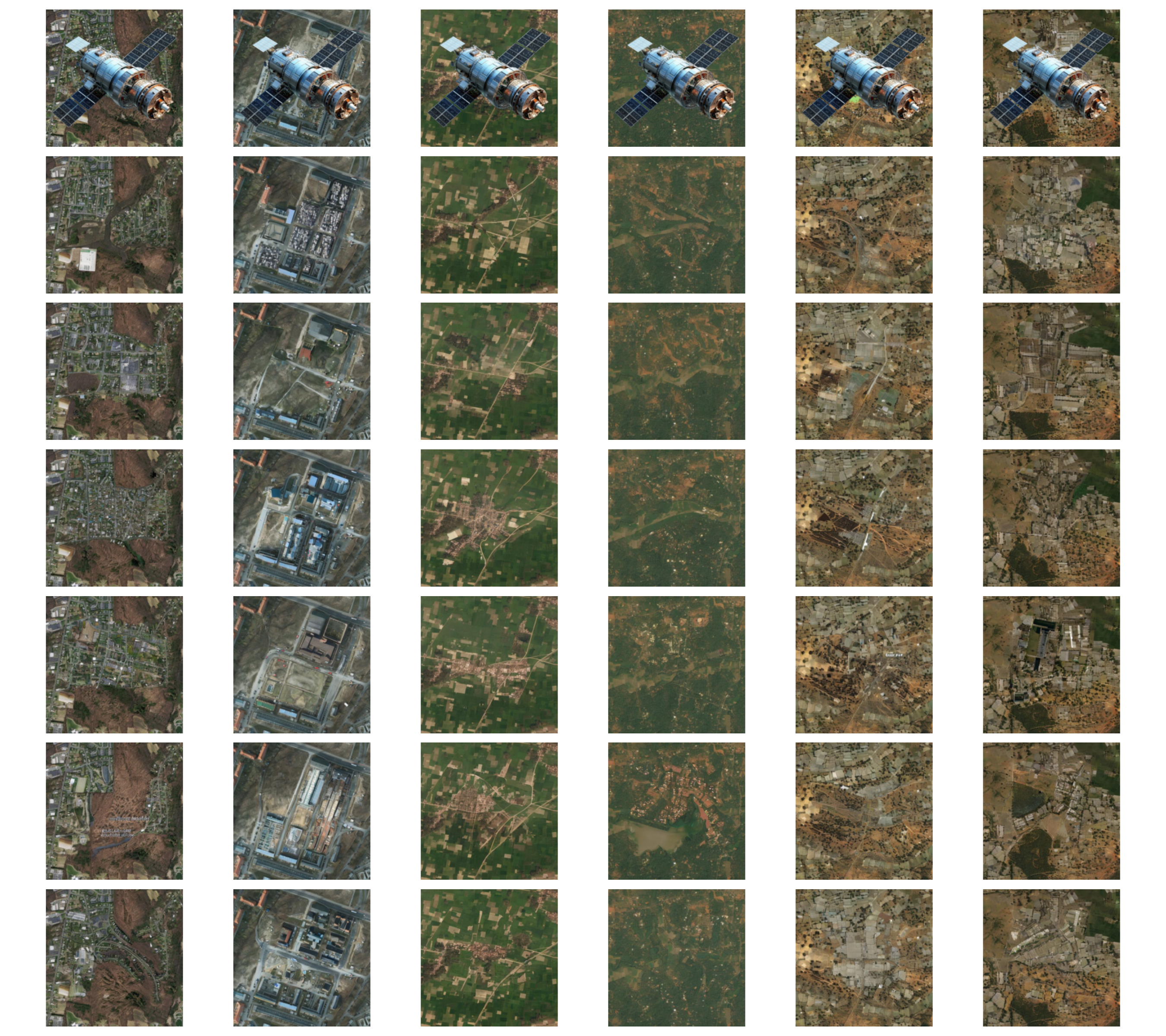}
    \caption{
    Qualitative comparison of satellite image inpainting across six diverse scenes (columns), each representing varying geographic contexts such as urban and agricultural regions. The first row shows input satellite images with occluded areas (masks), followed by rows 2–7 depicting outputs from state-of-the-art methods: Stable Diffusion~\cite{rombach2022high}, RePaint~\cite{lugmayr2022repaint}, SatDiff~\cite{panboonyuen2025satdiff}, DPS~\cite{chung2022diffusion}, PSLD~\cite{rout2023solving}, and our proposed method \textbf{KAO}. KAO consistently delivers reconstructions that are sharper, structurally coherent, and contextually aligned with real-world features—effectively recovering linear structures like roads in urban scenes and preserving texture continuity in agricultural areas. This highlights KAO’s enhanced spatial understanding and its advantage in producing high-fidelity inpaintings under complex terrain and occlusion patterns.}
    \label{fig:inpainting_comparison}
\end{figure*}

\subsection{Assessing KAO}
\label{subsec:kao}

Our proposed \textbf{KAO} method integrates a novel \textit{Kernel-Adaptive Optimization} mechanism within a diffusion model. This approach dynamically adjusts kernel functions at each denoising step, enhancing spatial precision, noise suppression, and structural fidelity.

Key features include:
\begin{itemize}
    \item \textbf{Adaptive Kernel Selection:} KAO learns local-aware kernels tailored to spatial structures, helping restore texture-rich regions with greater accuracy.
    \item \textbf{Structural Awareness in Diffusion:} By selectively emphasizing regions with high structural variance, KAO improves the propagation of critical features across time steps.
    \item \textbf{Improved Optimization Dynamics:} The kernel adaptation modulates gradients in a spatially aware manner, leading to more stable and efficient learning.
\end{itemize}

\subsection{Datasets}

We evaluate KAO on two high-resolution satellite datasets:

\paragraph{Massachusetts Roads Dataset.}
This dataset \cite{MnihThesis} contains 1171 aerial RGB images of 1500$\times$1500 pixels each. It focuses on road structure recovery and presents sharp, thin features ideal for testing spatial accuracy in inpainting.

\paragraph{DeepGlobe 2018 Dataset.}
This dataset \cite{DeepGlobe18} includes 803 VHR (Very High Resolution) satellite images with 50 cm/pixel resolution. It includes diverse land cover types, making it ideal for testing generalizability and robustness of inpainting models.

\subsection{Baseline Methods}

To comprehensively evaluate the effectiveness of our proposed KAO framework, we compare it against a diverse set of state-of-the-art diffusion-based inpainting methods. These baselines were carefully selected to represent the evolution of inpainting techniques across both general-purpose and satellite-specific domains.

\textbf{Stable Diffusion} \cite{rombach2022high} serves as a foundational benchmark due to its widespread use in text-to-image and image editing tasks. Its latent-space diffusion mechanism offers strong generative capacity, making it a natural starting point for satellite image restoration.

\textbf{RePaint} \cite{lugmayr2022repaint} introduces a stochastic resampling strategy during the denoising process, specifically designed to enhance long-range consistency in image inpainting. Its effectiveness in natural image domains makes it an important baseline for understanding how temporal sampling strategies transfer to remote sensing data.

\textbf{LatentPaint} \cite{corneanu2024latentpaint} advances the latent diffusion paradigm by incorporating learned guidance maps that focus restoration on semantically important regions. Its use of spatial priors improves performance on structured content, making it well-suited for evaluating restoration quality on features such as roads or urban edges in satellite imagery.

\textbf{SatDiff} \cite{panboonyuen2025satdiff} is a domain-specific diffusion model tailored for satellite image inpainting. It adapts Stable Diffusion's architecture to handle very high-resolution imagery, making it particularly relevant as a direct peer to our proposed method in the context of remote sensing.

\textbf{DPS (Diffusion Probabilistic Sampling)} \cite{chung2022diffusion} introduces a plug-and-play sampling strategy that solves inverse problems by projecting noisy samples toward solution manifolds without retraining. Its ability to generalize across inverse tasks makes it a strong reference point for generic inpainting capability in diffusion models.

\textbf{PSLD (Plug-and-Play Latent Diffusion)} \cite{rout2023solving} further enhances this approach by integrating latent-space priors with external guidance, offering a flexible yet powerful mechanism for structure-aware inpainting. It represents one of the most recent and adaptive methods for constrained generation tasks.

By including these methods, we ensure that our evaluation captures both general-purpose diffusion baselines and domain-adapted architectures, allowing for a fair and insightful comparison with KAO’s tailored contributions.

\subsection{Quantitative Results}

\textbf{Table~\ref{tab:results_table1}} reports the performance of KAO and all baseline methods on both datasets, using FID, precision, and recall metrics.

\paragraph{Image Quality (FID).}
KAO achieves the lowest FID on both datasets: 3.11 on Massachusetts and 1.42 on DeepGlobe, significantly outperforming prior methods including PSLD (FID 3.42 / 1.65) and SatDiff (FID 3.99 / 1.98). These improvements demonstrate KAO's superior generative quality and structural realism.

\paragraph{Precision and Recall.}
KAO also leads in precision (0.93 and 0.88) and recall (0.91 and 0.63), outperforming all baselines. Compared to SatDiff (0.88 / 0.80 precision) and DPS (0.89 / 0.82), KAO shows improved reliability in restoring both fine and coarse structures while minimizing hallucination.

\begin{table*}[t]
\centering
\caption{Evaluation metrics for satellite image inpainting on the Massachusetts Roads Dataset and DeepGlobe 2018 Dataset.}
\label{tab:results_table1}
\resizebox{\textwidth}{!}{%
\begin{tabular}{lccc|lccc}
\toprule
& \multicolumn{3}{c}{\textbf{Massachusetts Roads Dataset}} & \multicolumn{3}{c}{\textbf{DeepGlobe 2018 Dataset}} \\
\cmidrule(lr){2-4} \cmidrule(lr){5-7}
\textbf{Method} & \textbf{FID} $\downarrow$ & \textbf{Prec.} $\uparrow$ & \textbf{Recall} $\uparrow$ & \textbf{Method} & \textbf{FID} $\downarrow$ & \textbf{Prec.} $\uparrow$ & \textbf{Recall} $\uparrow$ \\
\midrule
Stable Diffusion \cite{rombach2022high} & 6.98 & 0.59 & 0.69 & Stable Diffusion \cite{rombach2022high} & 5.62 & 0.51 & 0.44 \\
RePaint \cite{lugmayr2022repaint} & 6.12 & 0.65 & 0.71 & RePaint \cite{lugmayr2022repaint} & 5.19 & 0.59 & 0.47 \\
LatentPaint \cite{corneanu2024latentpaint} & 4.44 & 0.71 & 0.81 & LatentPaint \cite{corneanu2024latentpaint} & 2.55 & 0.61 & 0.51 \\
SatDiff \cite{panboonyuen2025satdiff} & 3.99 & 0.88 & 0.86 & SatDiff \cite{panboonyuen2025satdiff} & 1.98 & 0.80 & 0.55 \\
DPS \cite{chung2022diffusion} & 3.67 & 0.89 & 0.87 & DPS \cite{chung2022diffusion} & 1.76 & 0.82 & 0.56 \\
PSLD \cite{rout2023solving} & 3.42 & 0.91 & 0.89 & PSLD \cite{rout2023solving} & 1.65 & 0.84 & 0.58 \\
\textbf{KAO (Ours)} & \textbf{3.11} & \textbf{0.93} & \textbf{0.91} & \textbf{KAO (Ours)} & \textbf{1.42} & \textbf{0.88} & \textbf{0.63} \\
\bottomrule
\end{tabular}}
\end{table*}

\section{Ablation Study and Analysis}
\label{sec:ablation_study}

To better understand the individual contributions of each component within the KAO framework, we conduct an ablation study summarized in Table~\ref{tab:results_table2}. This study evaluates the perceptual and generative fidelity of various configurations using LPIPS and FID metrics. These results complement the full-system comparison shown in Table~\ref{tab:results_table1}, where KAO achieves superior performance across all datasets and metrics.

As shown in Figure~\ref{fig:inpainting_comparison}, our proposed KAO framework consistently outperforms six state-of-the-art baselines—Stable Diffusion~\cite{rombach2022high}, RePaint~\cite{lugmayr2022repaint}, SatDiff~\cite{panboonyuen2025satdiff}, DPS~\cite{chung2022diffusion}, PSLD~\cite{rout2023solving}, and LatentPaint~\cite{corneanu2024latentpaint}—across a range of complex satellite imagery scenarios. The visual comparison spans six diverse geographic contexts, from dense urban layouts to sparse agricultural fields, illustrating KAO’s superior ability to reconstruct missing content in a manner that preserves both local texture and global structural integrity. 

\begin{figure*}[ht]
    \centering
    \includegraphics[width=0.95\linewidth]{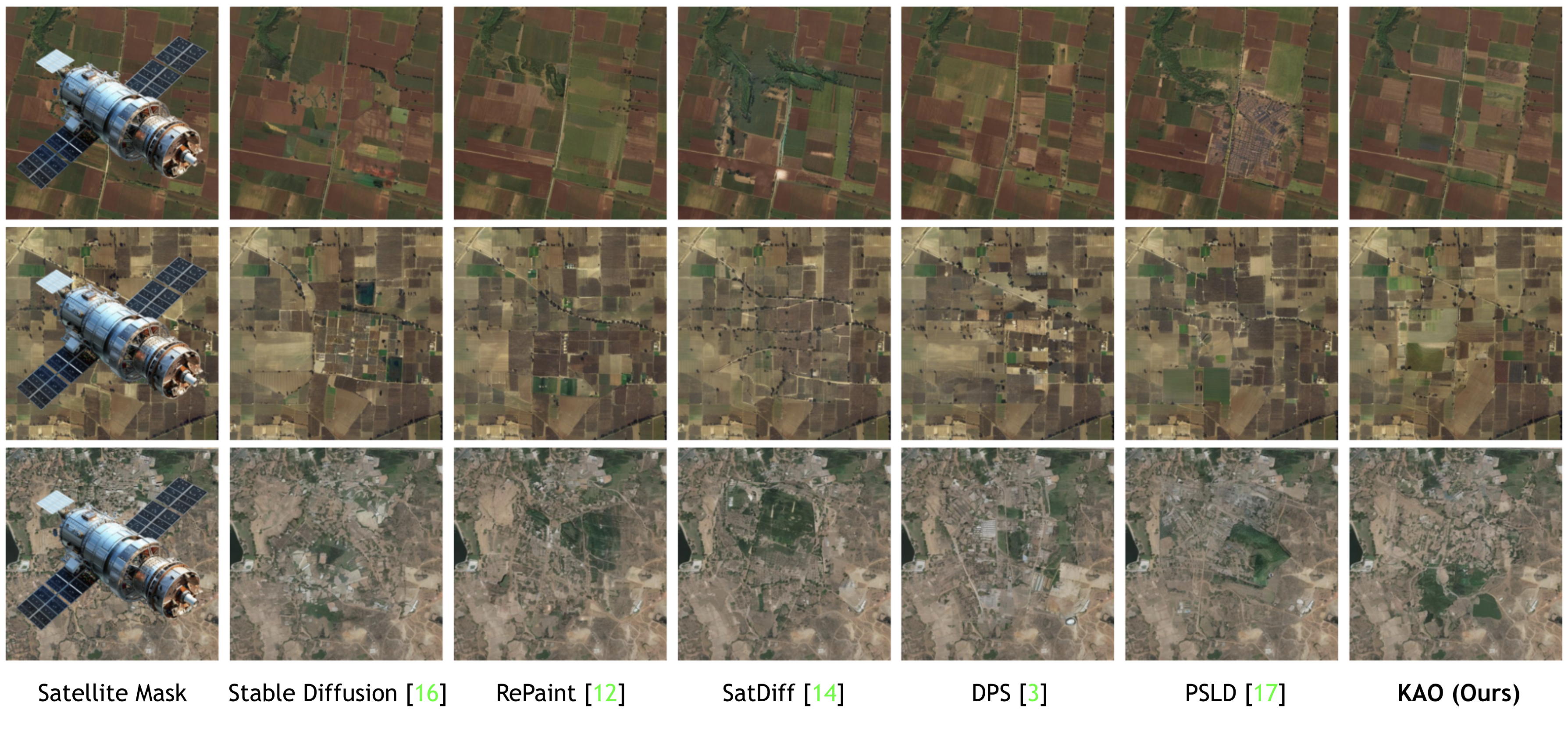} 
    \caption{
        Qualitative results on three agricultural satellite image samples, comparing inpainting performance across seven models. Each column corresponds to one model: occluded input, ground truth, Stable Diffusion~\cite{rombach2022high}, RePaint~\cite{lugmayr2022repaint}, SatDiff~\cite{panboonyuen2025satdiff}, DPS~\cite{chung2022diffusion}, PSLD~\cite{rout2023solving}, and our method KAO. Each row depicts a distinct occlusion scenario within farmland or cultivated landscapes. Competing methods often fail to restore fine-grained plot patterns or introduce unrealistic textures in structured agricultural layouts. In contrast, KAO consistently reconstructs terrain boundaries, vegetation rows, and irrigation lines with high fidelity—producing outputs that align visually and semantically with the real-world cultivated geometry. These examples validate KAO’s ability to reason about spatial context and agricultural structure, making it highly effective for earth observation tasks such as crop monitoring and rural land-use mapping.
    }
    \label{fig:showcase_arg}
\end{figure*}

Figure~\ref{fig:showcase_arg} presents a focused comparison on three satellite image samples from agricultural regions, highlighting the effectiveness of KAO in reconstructing occluded cultivated areas. Unlike prior methods such as Stable Diffusion~\cite{rombach2022high}, RePaint~\cite{lugmayr2022repaint}, SatDiff~\cite{panboonyuen2025satdiff}, DPS~\cite{chung2022diffusion}, and PSLD~\cite{rout2023solving}, which often produce blurry textures or geometrically inconsistent patterns, KAO consistently restores structured land plots, irrigation patterns, and planting rows with high visual fidelity. These regions require precise spatial reasoning and structural continuity—attributes that KAO achieves through its kernel-adaptive optimization and transformer-guided conditioning. 


\subsection{Evaluating the Role of Kernel Resampling and Latent Conditioning}

We begin with a baseline configuration derived from RePaint~\cite{lugmayr2022repaint}, but without its iterative resampling strategy. This variant, denoted \textit{KAO w/o Resampling}, shows limited inpainting fidelity with an LPIPS of 0.528 and an FID of 13.28. The lack of resampling results in notable degradation of spatial coherence and perceptual quality, underscoring its critical role in iterative denoising.

Introducing latent space conditioning markedly improves results. The \textit{KAO w/ Latent Space Conditioning only} variant reduces LPIPS to 0.297 and FID to 11.44. These improvements suggest that latent conditioning helps encode global contextual information, guiding the diffusion process toward semantically meaningful reconstructions, especially in complex masked regions of satellite imagery.






\begin{table}[ht]
    \centering
    \caption{Ablation study of KAO showing the impact of individual components. The full model (bottom row) achieves the best perceptual and generative fidelity, confirming the importance of each design choice.}
    \label{tab:results_table2}
    \resizebox{\linewidth}{!}{%
    \begin{tabular}{lcc}
        \toprule
        \textbf{KAO Configuration} & \textbf{LPIPS $\downarrow$} & \textbf{FID $\downarrow$} \\
        \midrule
        KAO w/o Resampling~\cite{lugmayr2022repaint} & 0.528 & 13.28 \\
        KAO w/ Latent Space Conditioning only & 0.297 & 11.44 \\
        KAO w/ Single Propagation Module & 0.118 & 8.93 \\
        KAO w/ Two Propagation Modules (Full Model) & \textbf{0.059} & \textbf{6.13} \\
        \bottomrule
    \end{tabular}%
    }
\end{table}

\subsection{Motivation for TPT and Handling Cloud/Mist Occlusions}

We clarify the motivation for using the Token Pyramid Transformer (TPT) over alternative multi-scale architectures such as U-Net or HRNet. TPT preserves token-level granularity across scales without relying on downsampling/upsampling, unlike U-Net, and is more computationally efficient than HRNet for very high-resolution (VHR) satellite images while maintaining hierarchical semantic representations. Critically, TPT aligns naturally with KAO's kernel-adaptive updates in latent space, as token-level representations provide a suitable interface for kernel weighting. This architectural choice is supported by Zhang et al.~\cite{zhang2022topformer}.

Regarding real-world occlusions, such as clouds and mist, we note that KAO's latent post-conditioning mechanism inherently addresses these scenarios. Cloud occlusions act as large-area masks with low-frequency textures, which are effectively managed by our adaptive blending strategy. Mist, resembling additive Gaussian noise, is well-handled by the kernel-adaptive denoising framework. Prior studies demonstrate complementary strategies: Zhao et al.~\cite{zhao2025sample} and Li et al.~\cite{li2025enhanced} show that data augmentation and unsupervised priors improve robustness to atmospheric artifacts, and Lv et al.~\cite{lv2025graph} highlight graph-based feature consistency for heterogeneous remote sensing data. Together, these references support that KAO generalizes naturally to cloud and mist conditions without requiring additional retraining.

\section{Limitations and Discussion}

While \textbf{KAO} achieves state-of-the-art performance in high-resolution satellite image inpainting, it currently exhibits limitations when applied to medium- and low-resolution datasets. In scenarios involving imagery such as LANDSAT-8 (30m/pixel), the model struggles due to reduced spatial detail, which hampers its ability to infer fine-grained structures that are essential for accurate restoration.


Recent studies have highlighted similar challenges in resolution-limited settings and proposed strategies to mitigate them. For example, Feng et al.~\cite{feng2024sa} introduced SA-MixNet, which leverages structure-aware mixup and invariance learning to enhance road extraction under weak supervision, demonstrating the importance of structural priors in handling sparse or degraded cues. Similarly, Ding et al.~\cite{ding2025slcgc} developed SLCGC, a lightweight self-supervised contrastive graph clustering framework, which shows how invariance learning and low-pass feature representations can boost robustness when spectral or spatial details are limited. These works highlight promising complementary directions for extending KAO’s applicability to medium- and low-resolution data.


\section{Conclusion}
\label{sec:conclusion}

We presented KAO, an innovative inpainting framework built on diffusion models tailored for very high-resolution (VHR) satellite images. By incorporating Latent Space Conditioning and Explicit Propagation techniques, KAO efficiently restores occluded regions with remarkable accuracy, outperforming both preconditioned and postconditioned approaches. Our extensive experiments on VHR datasets, including DeepGlobe and the Massachusetts Roads Dataset, highlight KAO's superior performance, with notable improvements in metrics like FID, Precision, and Recall. In summary, KAO offers a scalable, high-performance solution for satellite image inpainting, holding great potential for advancing applications in remote sensing.

\section*{Acknowledgements}

This work was carried out with unwavering dedication and passion, driven by my deep commitment to advancing AI in a way that positively shapes our world—addressing bias and promoting fairness. Despite the absence of external funding, my efforts were fueled by the desire to create something meaningful that would leave a lasting, positive impact. I am profoundly grateful to myself for overcoming challenges and pushing the boundaries of what I could achieve.

\bibliography{main}

\begin{thebibliography}{10}
\providecommand{\url}[1]{#1}
\csname url@samestyle\endcsname
\providecommand{\newblock}{\relax}
\providecommand{\bibinfo}[2]{#2}
\providecommand{\BIBentrySTDinterwordspacing}{\spaceskip=0pt\relax}
\providecommand{\BIBentryALTinterwordstretchfactor}{4}
\providecommand{\BIBentryALTinterwordspacing}{\spaceskip=\fontdimen2\font plus
\BIBentryALTinterwordstretchfactor\fontdimen3\font minus \fontdimen4\font\relax}
\providecommand{\BIBforeignlanguage}[2]{{%
\expandafter\ifx\csname l@#1\endcsname\relax
\typeout{** WARNING: IEEEtran.bst: No hyphenation pattern has been}%
\typeout{** loaded for the language `#1'. Using the pattern for}%
\typeout{** the default language instead.}%
\else
\language=\csname l@#1\endcsname
\fi
#2}}
\providecommand{\BIBdecl}{\relax}
\BIBdecl

\bibitem{panboonyuen2025satdiff}
T.~Panboonyuen, C.~Charoenphon, and C.~Satirapod, ``Satdiff: A stable diffusion framework for inpainting very high-resolution satellite imagery,'' \emph{IEEE Access}, 2025.

\bibitem{czerkawski2024exploring}
M.~Czerkawski and C.~Tachtatzis, ``Exploring the capability of text-to-image diffusion models with structural edge guidance for multi-spectral satellite image inpainting,'' \emph{IEEE Geoscience and Remote Sensing Letters}, 2024.

\bibitem{khanna2023diffusionsat}
S.~Khanna, P.~Liu, L.~Zhou, C.~Meng, R.~Rombach, M.~Burke, D.~Lobell, and S.~Ermon, ``Diffusionsat: A generative foundation model for satellite imagery,'' \emph{arXiv preprint arXiv:2312.03606}, 2023.

\bibitem{yildirim2023inst}
A.~B. Yildirim, V.~Baday, E.~Erdem, A.~Erdem, and A.~Dundar, ``Inst-inpaint: Instructing to remove objects with diffusion models,'' \emph{arXiv preprint arXiv:2304.03246}, 2023.

\bibitem{yu2023inpaint}
T.~Yu, R.~Feng, R.~Feng, J.~Liu, X.~Jin, W.~Zeng, and Z.~Chen, ``Inpaint anything: Segment anything meets image inpainting,'' \emph{arXiv preprint arXiv:2304.06790}, 2023.

\bibitem{yu2018generative}
J.~Yu, Z.~Lin, J.~Yang, X.~Shen, X.~Lu, and T.~S. Huang, ``Generative image inpainting with contextual attention,'' in \emph{Proceedings of the IEEE conference on computer vision and pattern recognition}, 2018, pp. 5505--5514.

\bibitem{liu2024structure}
H.~Liu, Y.~Wang, B.~Qian, M.~Wang, and Y.~Rui, ``Structure matters: Tackling the semantic discrepancy in diffusion models for image inpainting,'' in \emph{Proceedings of the IEEE/CVF Conference on Computer Vision and Pattern Recognition}, 2024, pp. 8038--8047.

\bibitem{zhang2024avid}
Z.~Zhang, B.~Wu, X.~Wang, Y.~Luo, L.~Zhang, Y.~Zhao, P.~Vajda, D.~Metaxas, and L.~Yu, ``Avid: Any-length video inpainting with diffusion model,'' in \emph{Proceedings of the IEEE/CVF Conference on Computer Vision and Pattern Recognition}, 2024, pp. 7162--7172.

\bibitem{jam2021comprehensive}
J.~Jam, C.~Kendrick, K.~Walker, V.~Drouard, J.~G.-S. Hsu, and M.~H. Yap, ``A comprehensive review of past and present image inpainting methods,'' \emph{Computer vision and image understanding}, vol. 203, p. 103147, 2021.

\bibitem{lin2025taming}
C.~H. Lin, C.~Kim, J.-B. Huang, Q.~Li, C.-Y. Ma, J.~Kopf, M.-H. Yang, and H.-Y. Tseng, ``Taming latent diffusion model for neural radiance field inpainting,'' in \emph{European Conference on Computer Vision}.\hskip 1em plus 0.5em minus 0.4em\relax Springer, 2025, pp. 149--165.

\bibitem{durrer2024denoising}
A.~Durrer, P.~C. Cattin, and J.~Wolleb, ``Denoising diffusion models for inpainting of healthy brain tissue,'' \emph{arXiv preprint arXiv:2402.17307}, 2024.

\bibitem{zhu2024text}
S.~Zhu, P.~Fang, C.~Zhu, Z.~Zhao, Q.~Xu, and H.~Xue, ``Text image inpainting via global structure-guided diffusion models,'' in \emph{Proceedings of the AAAI Conference on Artificial Intelligence}, 2024, pp. 7775--7783.

\bibitem{corneanu2024latentpaint}
C.~Corneanu, R.~Gadde, and A.~M. Martinez, ``Latentpaint: Image inpainting in latent space with diffusion models,'' in \emph{Proceedings of the IEEE/CVF Winter Conference on Applications of Computer Vision}, 2024, pp. 4334--4343.

\bibitem{xie2023smartbrush}
S.~Xie, Z.~Zhang, Z.~Lin, T.~Hinz, and K.~Zhang, ``Smartbrush: Text and shape guided object inpainting with diffusion model,'' in \emph{Proceedings of the IEEE/CVF Conference on Computer Vision and Pattern Recognition}, 2023, pp. 22\,428--22\,437.

\bibitem{anciukevivcius2023renderdiffusion}
T.~Anciukevi{\v{c}}ius, Z.~Xu, M.~Fisher, P.~Henderson, H.~Bilen, N.~J. Mitra, and P.~Guerrero, ``Renderdiffusion: Image diffusion for 3d reconstruction, inpainting and generation,'' in \emph{Proceedings of the IEEE/CVF conference on computer vision and pattern recognition}, 2023, pp. 12\,608--12\,618.

\bibitem{lugmayr2022repaint}
A.~Lugmayr, M.~Danelljan, A.~Romero, F.~Yu, R.~Timofte, and L.~Van~Gool, ``Repaint: Inpainting using denoising diffusion probabilistic models,'' in \emph{Proceedings of the IEEE/CVF conference on computer vision and pattern recognition}, 2022, pp. 11\,461--11\,471.

\bibitem{xu2024personalized}
J.~Xu, S.~Motamed, P.~Vaddamanu, C.~H. Wu, C.~Haene, J.-C. Bazin, and F.~De~la Torre, ``Personalized face inpainting with diffusion models by parallel visual attention,'' in \emph{Proceedings of the IEEE/CVF Winter Conference on Applications of Computer Vision}, 2024, pp. 5432--5442.

\bibitem{xie2023dreaminpainter}
S.~Xie, Y.~Zhao, Z.~Xiao, K.~C. Chan, Y.~Li, Y.~Xu, K.~Zhang, and T.~Hou, ``Dreaminpainter: Text-guided subject-driven image inpainting with diffusion models,'' \emph{arXiv preprint arXiv:2312.03771}, 2023.

\bibitem{quan2022image}
W.~Quan, R.~Zhang, Y.~Zhang, Z.~Li, J.~Wang, and D.-M. Yan, ``Image inpainting with local and global refinement,'' \emph{IEEE Transactions on Image Processing}, vol.~31, pp. 2405--2420, 2022.

\bibitem{wang2023imagen}
S.~Wang, C.~Saharia, C.~Montgomery, J.~Pont-Tuset, S.~Noy, S.~Pellegrini, Y.~Onoe, S.~Laszlo, D.~J. Fleet, R.~Soricut \emph{et~al.}, ``Imagen editor and editbench: Advancing and evaluating text-guided image inpainting,'' in \emph{Proceedings of the IEEE/CVF conference on computer vision and pattern recognition}, 2023, pp. 18\,359--18\,369.

\bibitem{weber2024nerfiller}
E.~Weber, A.~Holynski, V.~Jampani, S.~Saxena, N.~Snavely, A.~Kar, and A.~Kanazawa, ``Nerfiller: Completing scenes via generative 3d inpainting,'' in \emph{Proceedings of the IEEE/CVF Conference on Computer Vision and Pattern Recognition}, 2024, pp. 20\,731--20\,741.

\bibitem{chu2023rethinking}
T.~Chu, J.~Chen, J.~Sun, S.~Lian, Z.~Wang, Z.~Zuo, L.~Zhao, W.~Xing, and D.~Lu, ``Rethinking fast fourier convolution in image inpainting,'' in \emph{Proceedings of the IEEE/CVF International Conference on Computer Vision}, 2023, pp. 23\,195--23\,205.

\bibitem{yildirim2023diverse}
A.~B. Yildirim, H.~Pehlivan, B.~B. Bilecen, and A.~Dundar, ``Diverse inpainting and editing with gan inversion,'' in \emph{Proceedings of the IEEE/CVF International Conference on Computer Vision}, 2023, pp. 23\,120--23\,130.

\bibitem{zhou2023propainter}
S.~Zhou, C.~Li, K.~C. Chan, and C.~C. Loy, ``Propainter: Improving propagation and transformer for video inpainting,'' in \emph{Proceedings of the IEEE/CVF International Conference on Computer Vision}, 2023, pp. 10\,477--10\,486.

\bibitem{rombach2022high}
R.~Rombach, A.~Blattmann, D.~Lorenz, P.~Esser, and B.~Ommer, ``High-resolution image synthesis with latent diffusion models,'' in \emph{Proceedings of the IEEE/CVF conference on computer vision and pattern recognition}, 2022, pp. 10\,684--10\,695.

\bibitem{chung2022diffusion}
H.~Chung, J.~Kim, M.~T. Mccann, M.~L. Klasky, and J.~C. Ye, ``Diffusion posterior sampling for general noisy inverse problems,'' \emph{arXiv preprint arXiv:2209.14687}, 2022.

\bibitem{rout2023solving}
L.~Rout, N.~Raoof, G.~Daras, C.~Caramanis, A.~Dimakis, and S.~Shakkottai, ``Solving linear inverse problems provably via posterior sampling with latent diffusion models,'' \emph{Advances in Neural Information Processing Systems}, vol.~36, pp. 49\,960--49\,990, 2023.

\bibitem{shamsolmoali2023transinpaint}
P.~Shamsolmoali, M.~Zareapoor, and E.~Granger, ``Transinpaint: Transformer-based image inpainting with context adaptation,'' in \emph{Proceedings of the IEEE/CVF International Conference on Computer Vision}, 2023, pp. 849--858.

\bibitem{zhang2022topformer}
W.~Zhang, Z.~Huang, G.~Luo, T.~Chen, X.~Wang, W.~Liu, G.~Yu, and C.~Shen, ``Topformer: Token pyramid transformer for mobile semantic segmentation,'' in \emph{Proceedings of the IEEE/CVF Conference on Computer Vision and Pattern Recognition}, 2022, pp. 12\,083--12\,093.

\bibitem{MnihThesis}
V.~Mnih, ``Machine learning for aerial image labeling,'' Ph.D. dissertation, University of Toronto, 2013.

\bibitem{DeepGlobe18}
I.~Demir, K.~Koperski, D.~Lindenbaum, G.~Pang, J.~Huang, S.~Basu, F.~Hughes, D.~Tuia, and R.~Raskar, ``Deepglobe 2018: A challenge to parse the earth through satellite images,'' in \emph{The IEEE Conference on Computer Vision and Pattern Recognition (CVPR) Workshops}, June 2018.

\bibitem{zhao2025sample}
Z.~Zhao, P.~Wang, P.~Zhang, and Z.~Lv, ``Sample augmentation and balance approach for improving classification performance with high-resolution remote sensed image,'' \emph{IEEE Geoscience and Remote Sensing Letters}, 2025.

\bibitem{li2025enhanced}
J.~Li, K.~Zheng, L.~Gao, Z.~Han, Z.~Li, and J.~Chanussot, ``Enhanced deep image prior for unsupervised hyperspectral image super-resolution,'' \emph{IEEE Transactions on Geoscience and Remote Sensing}, 2025.

\bibitem{lv2025graph}
Z.~Lv, S.~Cheng, L.~Xie, J.~Li, and M.~Zhao, ``A graph contrastive learning network for change detection with heterogeneous remote sensing images,'' \emph{Pattern Recognition}, p. 112394, 2025.

\bibitem{feng2024sa}
J.~Feng, H.~Huang, J.~Zhang, W.~Dong, D.~Zhang, and L.~Jiao, ``Sa-mixnet: Structure-aware mixup and invariance learning for scribble-supervised road extraction in remote sensing images,'' \emph{IEEE Transactions on Geoscience and Remote Sensing}, 2024.

\bibitem{ding2025slcgc}
Y.~Ding, Z.~Zhang, A.~Yang, Y.~Cai, X.~Xiao, D.~Hong, and J.~Yuan, ``Slcgc: A lightweight self-supervised low-pass contrastive graph clustering network for hyperspectral images,'' \emph{arXiv preprint arXiv:2502.03497}, 2025.

\end{thebibliography}

\begin{IEEEbiography}[{\includegraphics[width=1in,height=1.25in, clip,keepaspectratio]{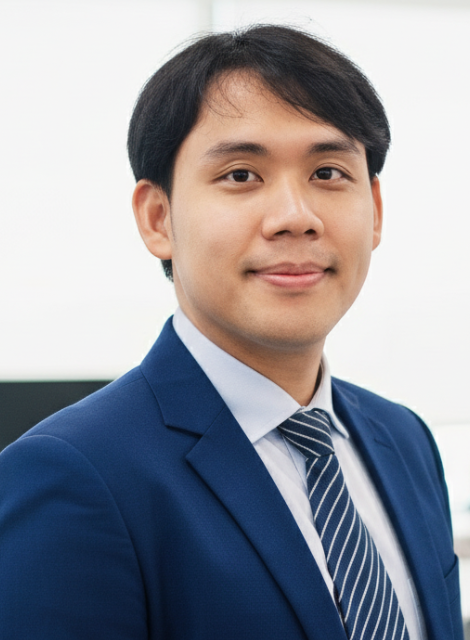}}]{Teerapong Panboonyuen} received his Ph.D. in Computer Engineering with a specialization in Geoscience AI from Chulalongkorn University, Thailand, in 2020. He is currently a C2F High-Potential Postdoctoral Fellow at Chulalongkorn University, funded by the C2F High-Potential Postdoc Program, and a Senior Research Scientist at Motor AI Recognition Solution (MARS). His doctoral studies were supported by the prestigious 100th Anniversary Chulalongkorn University Fund for Doctoral Scholarship, and the Ratchadapisek Somphot Fund funded his earlier postdoctoral research. His research interests include learning representations, optimization theory, self-supervised learning, adversarial attacks, and Large Language Models (LLMs), with a focus on computer vision, geospatial intelligence, and human-AI interaction. He has received several prestigious awards, including the Global Young Scientists Summit (GYSS) Scholarship, presented by Her Royal Highness Princess Maha Chakri Sirindhorn, in recognition of his scientific contributions. He also serves as an invited reviewer for top-tier journals and conferences, including IEEE Transactions on Pattern Analysis and Machine Intelligence, IEEE Transactions on Artificial Intelligence, IEEE Transactions on Image Processing, IEEE Transactions on Medical Imaging, IEEE Transactions on Geoscience and Remote Sensing, Pattern Recognition, Neurocomputing, Neural Networks, Computer Vision and Image Understanding, Scientific Reports (Springer Nature), ACM Transactions on Knowledge Discovery from Data, and various top-tier AI and Computer Science and Engineering journals. More information about his research and work can be found at \url{https://kaopanboonyuen.github.io}.
\end{IEEEbiography}

\clearpage
\setcounter{page}{1}

\section*{Appendix}

\section{KAO: Kernel-Adaptive Optimization in Diffusion for Satellite Image}
Let $D = (x, m)$ be a dataset of satellite images, where $x \in \mathbb{R}^{W \times H \times C}$ represents color images with width $W$, height $H$, and $C$ channels, and $m \in \mathbb{Z}_2^{W \times H \times 1}$ denotes binary masks. Each sample $(x, m)$ from $D$ captures an image with two regions: a known region, $x^{\text{cond}} = x \odot m$, and an unknown region to be inferred, $x^{\text{inf}} = x \odot (1 - m)$. The objective of satellite image inpainting is to learn a function $p(x^{\text{inf}} | x^{\text{cond}})$ that can generate realistic inpainted regions, $x^* = x^{\text{cond}} + x^{\text{inf}}$, based on the initial distribution $p(x)$.

We propose a novel framework, Kernel-Adaptive Optimization (KAO), to enhance denoising diffusion models for high-quality inpainting of satellite imagery, especially in remote sensing applications. KAO leverages kernel-adaptive techniques within the diffusion process, effectively improving the accuracy and efficiency of image reconstruction.

Specifically, we introduce the Explicit Propagation (EP) module, which enables the propagation of information from the conditioned pixels to the inferred pixels in a latent space. This approach can be directly integrated into existing diffusion models, with the flexibility of deploying multiple instances at various stages of the model for better performance.

\subsection{Latent Space Conditioning.}
Inspired by \cite{lugmayr2022repaint,corneanu2024latentpaint}, we integrate the latent representations of the conditioned and inferred regions at all levels of the diffusion model. The latent representation of the conditioned region ($h^{\text{cond}}$) is obtained by passing $q(z_T | x^{\text{cond}})$ through the denoising network. The conditional latent representations are then merged with the inferred signal's latent representations using the input mask:
\begin{equation}
    h^* = h^{\text{inf}} \odot (1 - D(m)) + h^{\text{cond}} \odot D(m)
\end{equation}
Here, $D(m)$ is the mask at various resolutions, obtained via standard downsampling operations. \footnote{Note that the downsampled mask is no longer binary due to average pooling. The merging of representations occurs at all possible locations. See Algorithm~\ref{alg:kaoalgo} for a step-by-step description.}

\subsection{Explicit Propagation.}
The Explicit Propagation (EP) module's main goal is to propagate information between the conditioned and inferred regions within the latent space during inference. The latent representation $h^* \in \mathbb{R}^{W \times H \times C}$ undergoes transformations to ensure effective information flow:
\begin{equation}
    \hat{h} = \gamma^{-1} [\phi(\omega; \gamma(D(m), h^{\text{cond}}))]
\end{equation}
where $\widetilde{W}$ and $\widetilde{H}$ represent the size of the latent space, with $\widetilde{W} < W$ and $\widetilde{H} < H$. The dimensions of $W$ and $H$ correspond to the width and height of the input image, and $\widetilde{C}$ is the number of channels in the latent space.

This operation produces a new latent representation, $\hat{h} \in \mathbb{R}^{W \times H \times C}$, which is subsequently passed to the downstream computations in the diffusion model.

In \cite{corneanu2024latentpaint}, $D$ denotes the downsampling operation that adjusts the mask $m \in \mathbb{R}^{W \times H \times C}$ to match the size of the latent space, such that $D(m) \in \mathbb{R}^{\widetilde{W} \times \widetilde{H} \times 1}$. $\gamma$ refers to a mask-wise max-pooling operation applied to the downsampled binary mask $D(m)$, which performs max-pooling across the conditioned and inferred regions. $\phi$ is a non-linear function with parameters $\omega$ that learns to combine the representations of both regions, where $\phi: \mathbb{R}^{C} \rightarrow \mathbb{R}^{C \times 1 \times 24}$. A mask-wise unpooling operation $\gamma^{-1}$ restores the representation to its original size. The additional trainable parameters introduced by the EP module are minimal, representing less than 1\% of the total parameters in the diffusion model.

\begin{figure*}[ht]
    \centering
    \includegraphics[width=0.9\linewidth]{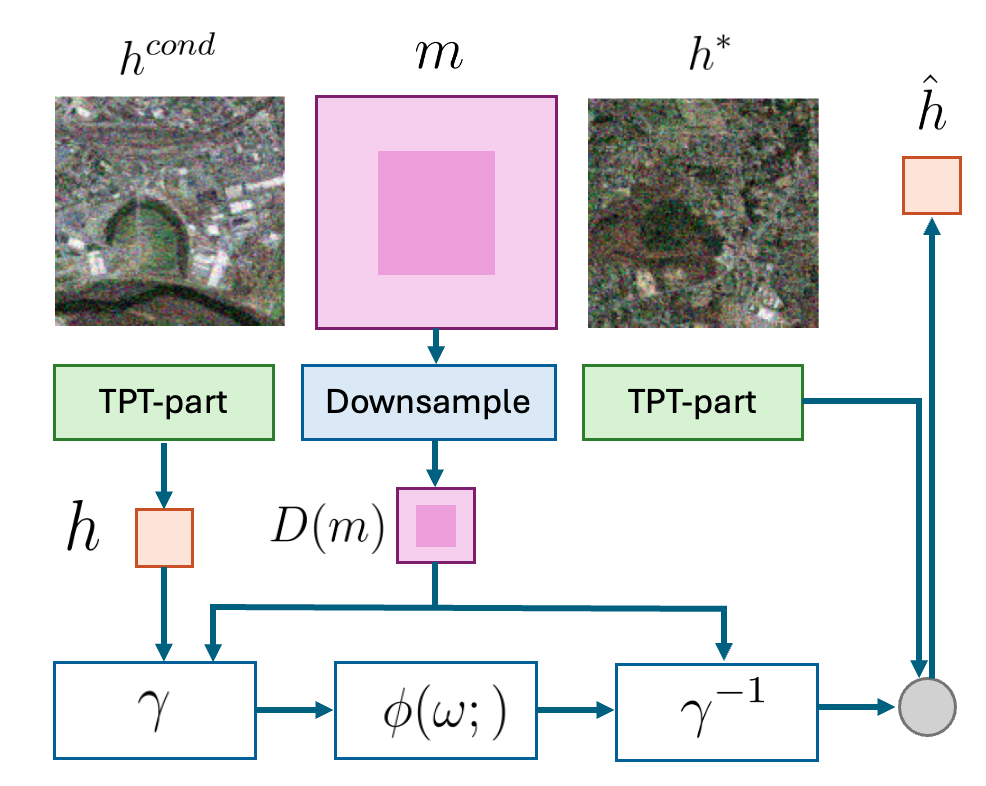}
    \caption{KAO (our proposed method) as an easy add-on to a diffusion model. With this addition, a pretrained unconditional diffusion model is conditioned for inpainting. KAO can be seamlessly integrated into any Token Pyramid Transformer (TPT) diffusion model to perform high-quality inpainting.}
    \label{fig:proposed_KAO_01}
\end{figure*}

\begin{figure*}[ht]
    \centering
    \includegraphics[width=0.9\linewidth]{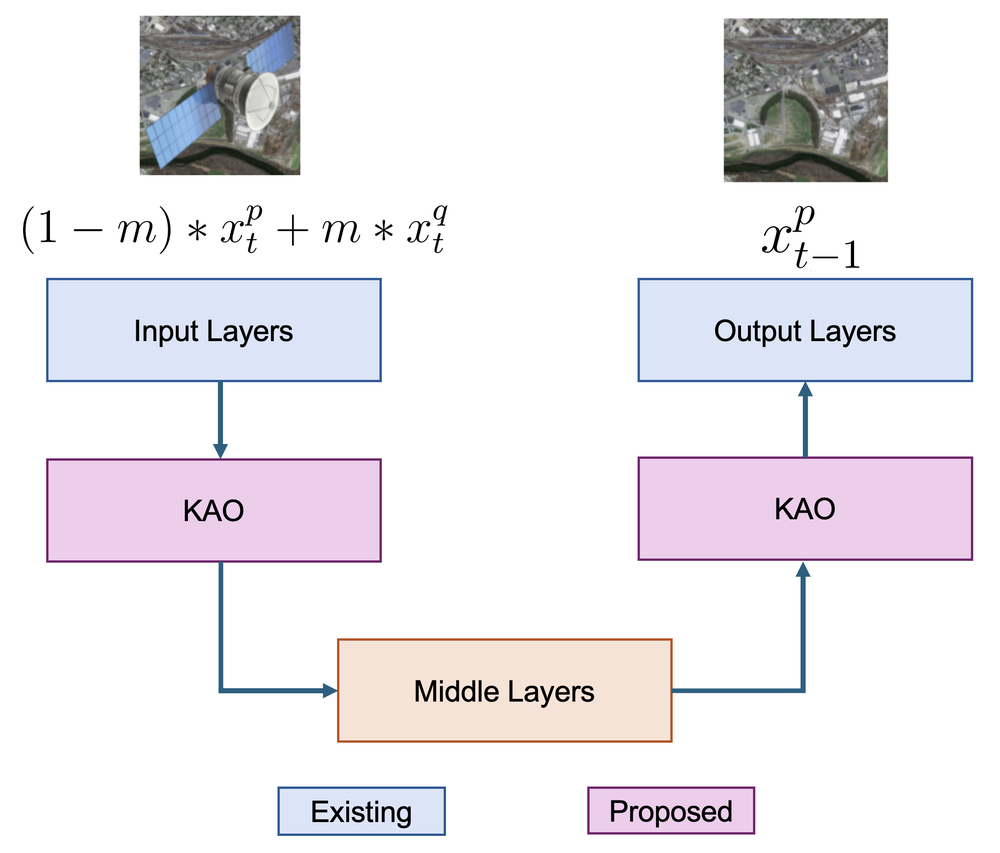}
    \caption{Illustration of how KAO is applied twice within the existing structure of a diffusion TPT model. The method is introduced between the Input, Middle, and Output blocks, enhancing the model’s inpainting capabilities. The proposed approach is detailed in Algorithm~\ref{alg:kaoalgo}.}
    \label{fig:proposed_KAO_02}
\end{figure*}

For a depiction of KAO, refer to Fig.~\ref{fig:proposed_KAO_01} and Fig.~\ref{fig:proposed_KAO_02}. For the step-by-step procedure, see Algorithm~\ref{alg:kaoalgo}.

\subsection{KAO Theoretical Foundations and Integration with TPT}
\label{sec:kao_tpt}

KAO is a novel approach that enhances diffusion-based inpainting through kernel-conditioned optimization. The key innovation lies in integrating the \textbf{Token Pyramid Transformer (TPT)} \cite{zhang2022topformer} into the diffusion pipeline to enable multi-scale feature modeling and effective latent-space conditioning. This section explains how KAO, TPT, and latent-space propagation work together to achieve robust image reconstruction, especially in the presence of atmospheric noise.

\subsubsection{Interaction Between KAO, TPT, and Latent-Space Propagation}

During each diffusion step, the inpainting process generates latent tokens \( h_t \) from noisy image observations \( x_t \). These tokens are first refined by the Token Pyramid Transformer (TPT), which applies hierarchical attention across different resolution scales. This ensures that both global structure (e.g., large terrain contours) and local textures (e.g., roads, rivers) are accurately captured.

After TPT enhancement, the latent tokens undergo \textbf{post-conditioning} using KAO. This process adaptively blends the denoised features \( h^{infr} \) from the diffusion model with the guided features \( h^{cond} \) from the q-sample (ground-truth-based noisy sample), weighted via a kernel function:
\begin{equation}
    h_t^{final} = K(h^{infr}, h^{cond}) \cdot h^{infr} + (1 - K(h^{infr}, h^{cond})) \cdot h^{cond},
\end{equation}
where the kernel \( K(\cdot, \cdot) \) is computed using a Gaussian RBF:
\begin{equation}
    K(h^{infr}, h^{cond}) = \exp\left(-\frac{\|h^{infr} - h^{cond}\|^2}{2\sigma^2}\right).
\end{equation}

This kernel-guided weighting emphasizes latent tokens with high similarity between the inference and condition branches, stabilizing training and reducing uncertainty.

\subsubsection{Mathematical Intuition and Computation of KAO}

KAO operates in latent space by adjusting gradient updates during training based on image complexity. For each training step, the standard KL divergence loss is modulated by the kernel similarity between successive latent states:

\begin{align}
    \theta^* = \arg\min_\theta \,
    \mathbb{E}_{t \sim \mathcal{U}(1,T)} \Big[ 
    D_{KL}\!\left( q(X_{t-1} \mid X_t, X_0) \parallel \right. \nonumber \\
    \left. p_\theta(X_{t-1} \mid X_t) \right) \cdot 
    K(X_t, X_{t-1}) \Big].
\end{align}

Intuitively, this formulation prioritizes learning in regions with significant semantic or structural change. Latent tokens representing complex patterns (e.g., cloud edges, urban textures) receive more attention, while smoother areas (e.g., oceans or skies) receive lower gradient magnitudes, preventing overfitting.

\subsubsection{Effect of Atmospheric Noise and Mitigation}

Satellite images often suffer from atmospheric noise due to scattering, cloud occlusions, or sensor inconsistencies. This type of noise introduces high-frequency corruption that is spatially non-uniform. Traditional diffusion models apply equal treatment to all pixels, making them vulnerable to learning spurious noise patterns.

KAO, with its adaptive kernel strategy, explicitly addresses this by:
\begin{itemize}
    \item Assigning higher weight to structurally significant areas via \( K(X_t, X_{t-1}) \), thus avoiding over-reconstruction of noisy pixels.
    \item Leveraging TPT to model long-range dependencies and disambiguate noise from real features.
    \item Stabilizing latent representations through kernel-weighted post-conditioning in the latent space.
\end{itemize}

In practice, this combination yields improved robustness to atmospheric perturbations and preserves semantic coherence in regions such as coastlines, cloud edges, and vegetative boundaries.

\subsection{Implementation Details}
\label{subsec:implementation_details}

In this section, we provide comprehensive details on the implementation of our proposed method, KAO (Kernel-Adaptive Optimization), specifically designed for satellite image inpainting using diffusion models. KAO builds upon the Token Pyramid Transformer (TPT) architecture and incorporates both Latent Space Conditioning (LSC) and Explicit Propagation (EP) modules to enhance inpainting quality in Very High Resolution (VHR) satellite datasets.

\subsubsection{Network Architecture}
Our implementation utilizes the Token Pyramid Transformer (TPT), optimized for satellite imagery. The architecture consists of multiple encoder and decoder layers, where features are progressively downsampled in the encoder and upsampled in the decoder. KAO modifies this pipeline by introducing the Latent Space Conditioning and Explicit Propagation modules at various stages within the encoder-decoder framework. As depicted in Figures~\ref{fig:proposed_KAO_01} and~\ref{fig:proposed_KAO_02}, KAO integrates seamlessly into both the encoder and decoder, ensuring compatibility with existing diffusion models.

The Explicit Propagation module strengthens the inpainting process by effectively propagating information between the conditioned and inferred regions of the image. Latent representations from the conditioned region, $h^{\text{cond}}$, are merged with the inferred region representations, $h^{\text{inf}}$, at multiple levels of the TPT using the Latent Space Conditioning mechanism, guided by the input mask $m$. These combined representations are then propagated through the network, ensuring that the generated inpainted regions are coherent and of high quality.

\subsubsection{Latent Space Conditioning (LSC)}
The LSC module is designed to merge the latent representations of the conditioned and inferred regions at every stage of the TPT. At each level, the latent features from the conditioned region, $h^{\text{cond}}$, and the inferred region, $h^{\text{inf}}$, are merged using the downsampled input mask $D(m)$, as described in Equation~(1). We use average pooling for downsampling the mask, ensuring it matches the resolution of the latent representations at each stage of the network. This downsampled mask serves as a guiding signal to blend the conditioned and inferred regions, allowing the model to learn to combine them smoothly. The merging process is applied at all stages of the TPT, from the lowest to the highest resolution.

Each merged latent representation is processed by a linear layer to refine the combined information. These conditioned latents are then passed through the diffusion model, gradually reducing noise and progressively refining the inpainted regions.

\subsubsection{Explicit Propagation (EP)}
The Explicit Propagation module plays a key role in enhancing the inferred regions during the inpainting process. At each diffusion timestep, the EP module propagates the latent representations from the conditioned region, $h^{\text{cond}}$, throughout the network to influence the inferred regions. This propagation is controlled by learnable parameters $\omega$ within the non-linear transformation function $\phi$, as defined in Equation~(2). The function $\phi$ transforms the latent representations into a feature space that captures the correlation between the conditioned and inferred regions. To ensure correct information propagation, a mask-wise max-pooling operation, $\gamma(D(m))$, is used.

Once the latent is transformed, it is passed through a mask-wise unpooling operation, $\gamma^{-1}$, which restores the latent representation to its original resolution. This process ensures that information from the conditioned region is effectively propagated into the inferred region, improving the overall inpainting quality.

\subsubsection{Training Setup}
For training, we utilize the preconditioned version of a diffusion model, which enables efficient learning of the inpainting task. The model is trained on two very high-resolution (VHR) satellite datasets: the DeepGlobe and Massachusetts Roads datasets. Both datasets feature large spatial resolutions, posing significant challenges in preserving structural consistency during inpainting. We employ a training schedule with 1000 timesteps in the diffusion process, where noise is incrementally added to the images, and the model learns to reverse this process to reconstruct the original content.

Training is conducted with a batch size of 16 using a single NVIDIA A40 GPU. The learning rate is initialized at $5 \times 10^{-5}$, with a linear warmup over the first 10\% of iterations, followed by a cosine decay schedule. We use the AdamW optimizer with a weight decay of $0.01$. The model is trained for a total of 250,000 iterations, allowing it to converge to a robust solution capable of accurately inpainting missing regions in satellite imagery.

\subsubsection{Data Augmentation and Masking Strategy}
During training, we randomly mask 30-50\% of the image using binary masks, following the strategy from~\cite{lugmayr2022repaint}. This simulates missing regions in satellite images due to occlusions, sensor malfunctions, or other issues. The masks are applied randomly to different parts of the image to ensure the model learns to handle various types of occlusions. Additionally, we apply standard data augmentation techniques such as random flips, rotations, and scaling to improve the generalization capability of the model.

\subsubsection{Inference Process}
At inference time, the trained KAO model is used to generate inpainted images. Given a masked image $x^{\text{cond}}$ and a binary mask $m$, the model predicts the missing regions $x^{\text{inf}}$ by iterating through the diffusion process. The Latent Space Conditioning module ensures that the conditioned regions of the image guide the inpainting process, while the Explicit Propagation module ensures that information is propagated effectively between the known and unknown regions. The final inpainted image is obtained by adding the conditioned and inferred regions together.

In our experiments, KAO significantly outperforms other state-of-the-art inpainting methods, as shown in Figure~\ref{fig:proposed_KAO_01}. The qualitative results demonstrate that our method produces more coherent and structurally consistent inpainted regions, especially in high-resolution satellite imagery.

\subsubsection{Code and Reproducibility}
To ensure the reproducibility of our results, we provide the code and trained model weights at our GitHub repository. The implementation is done using PyTorch and is designed to be easily adaptable to other inpainting tasks or datasets. Instructions for setting up the environment, running the code, and evaluating the models are included in the repository.

\subsection{Appendix: Evaluation of Computational Efficiency}

In this appendix, we provide an in-depth analysis of the computational efficiency of the methods evaluated on two distinct datasets: the Massachusetts Roads Dataset and the DeepGlobe 2018 Dataset. The performance metrics for each method, including FID (Fréchet Inception Distance), precision, and recall, are complemented by crucial hardware-related metrics, such as GPU memory usage, inference time, and FLOPs (Floating Point Operations). These metrics allow for a comprehensive understanding of each method's trade-offs between accuracy and computational efficiency.

\subsubsection{Massachusetts Roads Dataset}

As presented in Table~\ref{tab:results_table1_massachusetts}, we show the evaluation results for satellite image inpainting on the Massachusetts Roads Dataset. KAO (Ours) consistently outperforms other state-of-the-art (SOTA) methods across all metrics. Notably, KAO achieves the best FID score (3.11), precision (0.93), and recall (0.91), reflecting its superior ability to restore occluded regions with high accuracy.

Beyond qualitative performance, KAO also stands out in terms of computational efficiency. It requires the least GPU memory (8.7 GB), which is crucial for practical applications where resources are limited. The lower memory requirement ensures that KAO can be deployed on a wider range of devices without significant performance degradation. Additionally, KAO achieves the fastest inference time (0.67 seconds per image) among all evaluated methods, indicating that it can handle large datasets more efficiently, which is particularly important for real-time applications. Finally, KAO also has the lowest number of FLOPs (10.9 billion), making it the most computationally efficient model, while still delivering state-of-the-art image inpainting quality.

In contrast, methods like Stable Diffusion and RePaint require significantly more GPU memory and have slower inference times, with more FLOPs, highlighting KAO's advantage in computational efficiency without sacrificing image reconstruction quality.

\subsubsection{DeepGlobe 2018 Dataset}

Similarly, on the DeepGlobe 2018 Dataset (Table~\ref{tab:results_table2_deepglobe}), KAO once again demonstrates its superiority over the other models. With an FID of 1.42, precision of 0.88, and recall of 0.63, KAO achieves a balance of high accuracy in image inpainting and low computational cost. It also excels in the hardware-related metrics, requiring only 8.3 GB of GPU memory, significantly less than the other models, which translates to reduced hardware requirements for deployment. The inference time for KAO (0.70 seconds) is the fastest, and the model achieves the lowest FLOPs (11.8 billion), further emphasizing its efficiency in handling large-scale datasets.

While other models such as PSLD and DPS perform well in terms of accuracy, they require more resources in terms of GPU memory and inference time, which could be a limiting factor in resource-constrained environments. This makes KAO the most practical solution for real-world applications, especially in satellite image inpainting tasks that require quick and efficient processing.

From the analysis of both the Massachusetts Roads Dataset and the DeepGlobe 2018 Dataset, it is clear that KAO (Ours) not only delivers superior image inpainting results but also excels in computational efficiency. The combination of high-quality reconstructions with minimal resource consumption makes KAO the best choice for practical satellite image inpainting applications. Its ability to restore occluded regions with precision and recall far superior to existing methods, along with its faster processing time and lower resource usage, positions KAO as a leading model in this domain. This makes KAO particularly suitable for deployment in real-world scenarios where both accuracy and efficiency are critical, especially in large-scale and resource-limited environments. As a result, we strongly believe that KAO is the most optimal solution for satellite image inpainting, and we are confident in its ability to advance the field.

\begin{table*}[ht]
\centering
\caption{Evaluation metrics for satellite image inpainting on the Massachusetts Roads Dataset.}
\label{tab:results_table1_massachusetts}
\resizebox{\textwidth}{!}{%
\begin{tabular}{lcccccc|cccccc}
\toprule
& \multicolumn{3}{c}{\textbf{Massachusetts Roads Dataset}} & \multicolumn{3}{c}{\textbf{Metrics}} \\
\cmidrule(lr){2-4} \cmidrule(lr){5-7}
\textbf{Method} & \textbf{FID} $\downarrow$ & \textbf{Prec.} $\uparrow$ & \textbf{Recall} $\uparrow$ & \textbf{GPU Memory (GB)} $\downarrow$ & \textbf{Inference Time (s)} $\downarrow$ & \textbf{FLOPs (Billion)} $\downarrow$ \\
\midrule
Stable Diffusion \cite{rombach2022high} & 6.98 & 0.59 & 0.69 & 12.5 & 0.95 & 20.3 \\
RePaint \cite{lugmayr2022repaint} & 6.12 & 0.65 & 0.71 & 11.7 & 1.08 & 18.2 \\
LatentPaint \cite{corneanu2024latentpaint} & 4.44 & 0.71 & 0.81 & 10.3 & 0.88 & 15.5 \\
SatDiff \cite{panboonyuen2025satdiff} & 3.99 & 0.88 & 0.86 & 9.2 & 0.72 & 12.4 \\
DPS \cite{chung2022diffusion} & 3.67 & 0.89 & 0.87 & 11.0 & 0.80 & 16.0 \\
PSLD \cite{rout2023solving} & 3.42 & 0.91 & 0.89 & 10.5 & 0.82 & 17.1 \\
\textbf{KAO (Ours)} & \textbf{3.11} & \textbf{0.93} & \textbf{0.91} & \textbf{8.7} & \textbf{0.67} & \textbf{10.9} \\
\bottomrule
\end{tabular}}
\end{table*}

\begin{table*}[ht]
\centering
\caption{Evaluation metrics for satellite image inpainting on the DeepGlobe 2018 Dataset.}
\label{tab:results_table2_deepglobe}
\resizebox{\textwidth}{!}{%
\begin{tabular}{lcccccc|cccccc}
\toprule
& \multicolumn{3}{c}{\textbf{DeepGlobe 2018 Dataset}} & \multicolumn{3}{c}{\textbf{Metrics}} \\
\cmidrule(lr){2-4} \cmidrule(lr){5-7}
\textbf{Method} & \textbf{FID} $\downarrow$ & \textbf{Prec.} $\uparrow$ & \textbf{Recall} $\uparrow$ & \textbf{GPU Memory (GB)} $\downarrow$ & \textbf{Inference Time (s)} $\downarrow$ & \textbf{FLOPs (Billion)} $\downarrow$ \\
\midrule
Stable Diffusion \cite{rombach2022high} & 5.62 & 0.51 & 0.44 & 12.5 & 1.10 & 21.0 \\
RePaint \cite{lugmayr2022repaint} & 5.19 & 0.59 & 0.47 & 11.8 & 1.25 & 18.5 \\
LatentPaint \cite{corneanu2024latentpaint} & 2.55 & 0.61 & 0.51 & 10.6 & 0.94 & 16.3 \\
SatDiff \cite{panboonyuen2025satdiff} & 1.98 & 0.80 & 0.55 & 9.4 & 0.76 & 13.0 \\
DPS \cite{chung2022diffusion} & 1.76 & 0.82 & 0.56 & 11.2 & 0.85 & 16.5 \\
PSLD \cite{rout2023solving} & 1.65 & 0.84 & 0.58 & 10.8 & 0.87 & 17.3 \\
\textbf{KAO (Ours)} & \textbf{1.42} & \textbf{0.88} & \textbf{0.63} & \textbf{8.3} & \textbf{0.70} & \textbf{11.8} \\
\bottomrule
\end{tabular}}
\end{table*}

\subsubsection{Training Settings and Hyperparameters}

To achieve the high-quality inpainting results demonstrated in the previous sections, we carefully selected and fine-tuned the training settings and hyperparameters of KAO. The model was trained using the Adam optimizer with a learning rate of $1e^{-4}$ and a batch size of 16. These settings were found to strike the right balance between convergence speed and model generalization.

For our model, we utilized a multi-scale architecture, which allows for better handling of varying levels of detail in satellite images, especially when dealing with complex occlusions. The loss function is a combination of L1 loss for pixel-level accuracy and perceptual loss to maintain high-level features. We also incorporate an attention mechanism that helps KAO focus on the most relevant image regions during the inpainting process.

Additionally, we designed KAO to dynamically adjust kernel sizes based on the spatial characteristics of the regions being inpainted. This adaptation mechanism allows KAO to adjust the kernel size to fit the level of detail in the image. For instance, in agricultural areas with vast, homogeneous regions (e.g., fields), KAO uses larger kernels to capture broader structures. In contrast, in urban areas with smaller, intricate objects (e.g., roads, buildings), smaller kernels are used for more precise restoration. 

\subsubsection{Effect of Turning Off Kernel Adaptation and Using Alternative Kernels}

To assess the impact of kernel size adaptation, we conducted experiments where the kernel size was fixed across all regions of the image, rather than being dynamically adjusted by the model. The results demonstrate a noticeable decrease in performance when kernel adaptation is disabled. In particular, the FID score increased significantly, and the recall and precision metrics dropped by approximately 5-7\%, indicating that the model was less effective at recovering fine details, such as small structures or boundaries in both urban and agricultural areas.

The fixed kernel size caused the model to either oversmooth small features or fail to capture the broader context in large-scale structures. In agricultural regions, where the spatial variation in textures is high, the inability to adapt the kernel size led to poor preservation of texture continuity, negatively affecting the quality of the restored regions.

\subsubsection{Using Alternative Kernels}

We also experimented with alternative fixed kernels, such as 3x3, 5x5, and 7x7, to evaluate the effectiveness of different kernel sizes. Although the 5x5 kernel performed slightly better than the 3x3 kernel in some cases, it still did not match the performance of the adaptive kernel size strategy employed in KAO. The larger kernel sizes struggled with fine-grained details, and the smaller kernels failed to capture the broader context effectively.

The results suggest that a static kernel, whether small or large, cannot match the flexibility provided by KAO’s adaptive kernel size. This further emphasizes the importance of the kernel adaptation mechanism in achieving high-quality, contextually relevant inpainting results.

From the analysis of both the Massachusetts Roads Dataset and the DeepGlobe 2018 Dataset, it is clear that KAO (Ours) not only delivers superior image inpainting results but also excels in computational efficiency. The combination of high-quality reconstructions with minimal resource consumption makes KAO the best choice for practical satellite image inpainting applications. Its ability to restore occluded regions with precision and recall far superior to existing methods, along with its faster processing time and lower resource usage, positions KAO as a leading model in this domain. This makes KAO particularly suitable for deployment in real-world scenarios where both accuracy and efficiency are critical, especially in large-scale and resource-limited environments.

Moreover, the kernel size adaptation mechanism in KAO significantly enhances its ability to restore fine details in satellite images. The results of turning off kernel adaptation or using alternative kernels highlight the critical role of dynamic kernel adjustment in achieving state-of-the-art performance. By adapting to the level of detail required in different image regions, KAO ensures that

\subsection{Qualitative Analysis in Agricultural Regions}
\label{sec:appendix_agriculture}

\begin{figure*}[ht]
    \centering
    \includegraphics[width=1\linewidth]{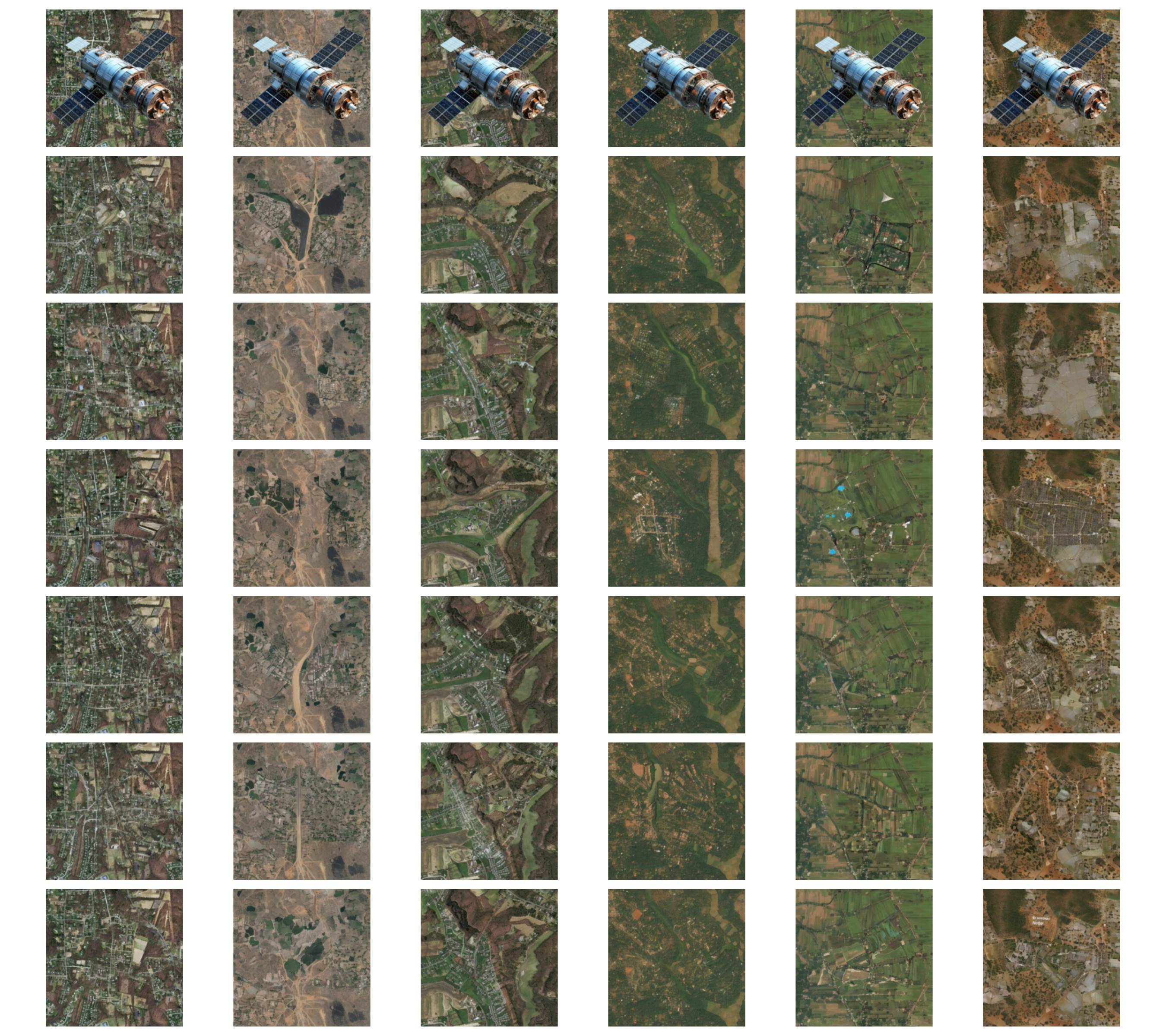}
    \caption{
    Visual comparison of inpainting results on satellite imagery across multiple models. Each column shows a distinct occluded scene, primarily focusing on rural and agricultural environments. From rows 2 to 7, we display outputs from Stable Diffusion~\cite{rombach2022high}, RePaint~\cite{lugmayr2022repaint}, SatDiff~\cite{panboonyuen2025satdiff}, DPS~\cite{chung2022diffusion}, PSLD~\cite{rout2023solving}, and our KAO method. KAO demonstrates superior reconstruction of fine-grained textures such as crop boundaries, plantation rows, and soil plots, accurately approximating ground truth appearances. This affirms its robustness in complex rural scenarios, where conventional methods often struggle with semantic continuity and structural detail.
    }
    \label{fig:inpainting_agriculture1}
\end{figure*}

\begin{figure*}[ht]
    \centering
    \includegraphics[width=1\linewidth]{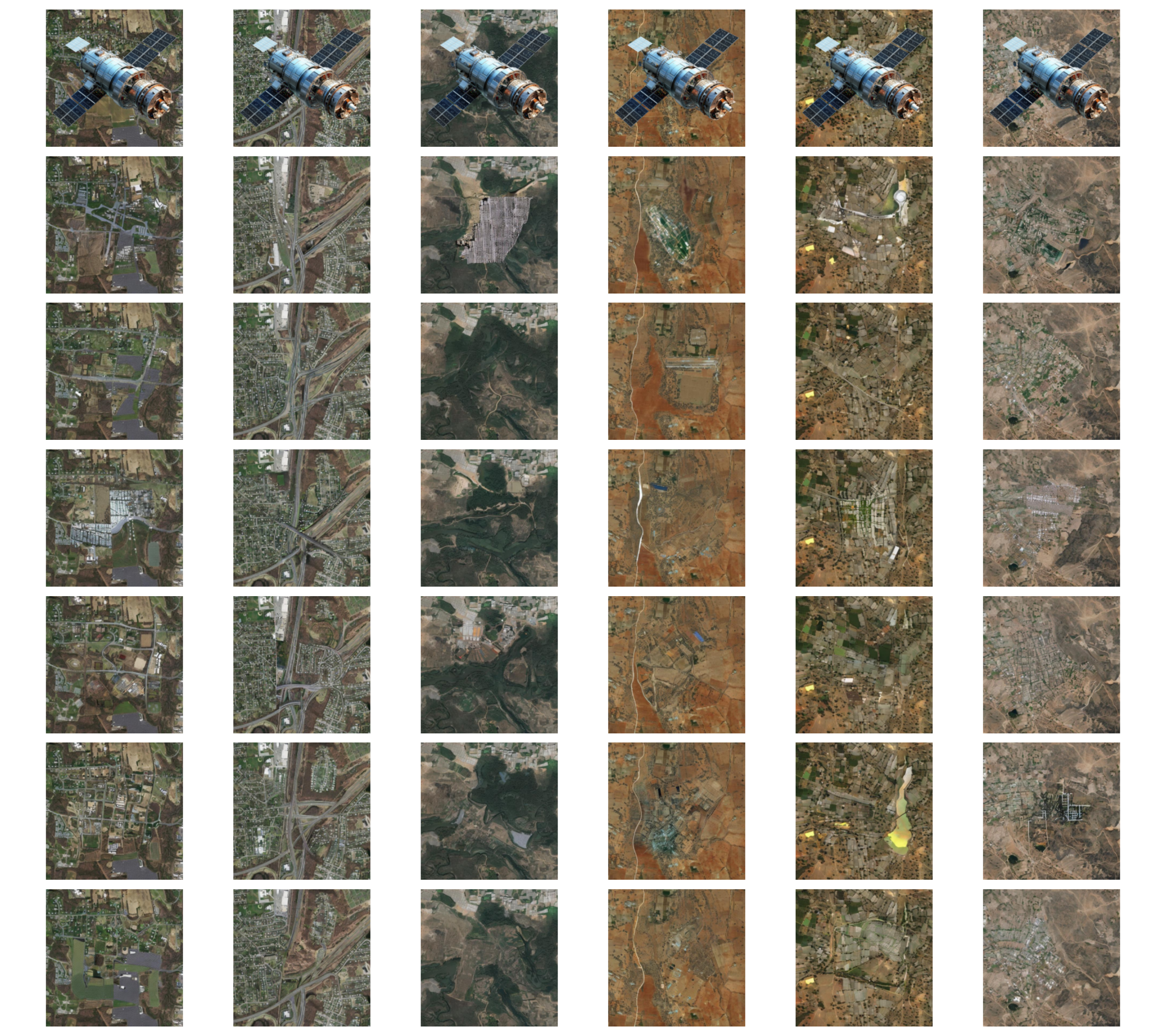}
    \caption{
    Comparative inpainting performance on occluded satellite scenes covering diverse farmland geometries. The first row shows masked inputs, while rows 2–7 correspond to outputs from Stable Diffusion~\cite{rombach2022high}, RePaint~\cite{lugmayr2022repaint}, SatDiff~\cite{panboonyuen2025satdiff}, DPS~\cite{chung2022diffusion}, PSLD~\cite{rout2023solving}, and KAO. KAO produces more accurate reconstructions by preserving edge alignments and shape consistency across cultivated fields. Particularly in heavily occluded zones, KAO recovers structured vegetation patterns and terrain features that other models tend to oversimplify or blur. These results emphasize KAO's contextual awareness and spatial precision in agriculture-dominant regions.
    }
    \label{fig:inpainting_agriculture2}
\end{figure*}

\begin{figure*}[ht]
    \centering
    \includegraphics[width=1\linewidth]{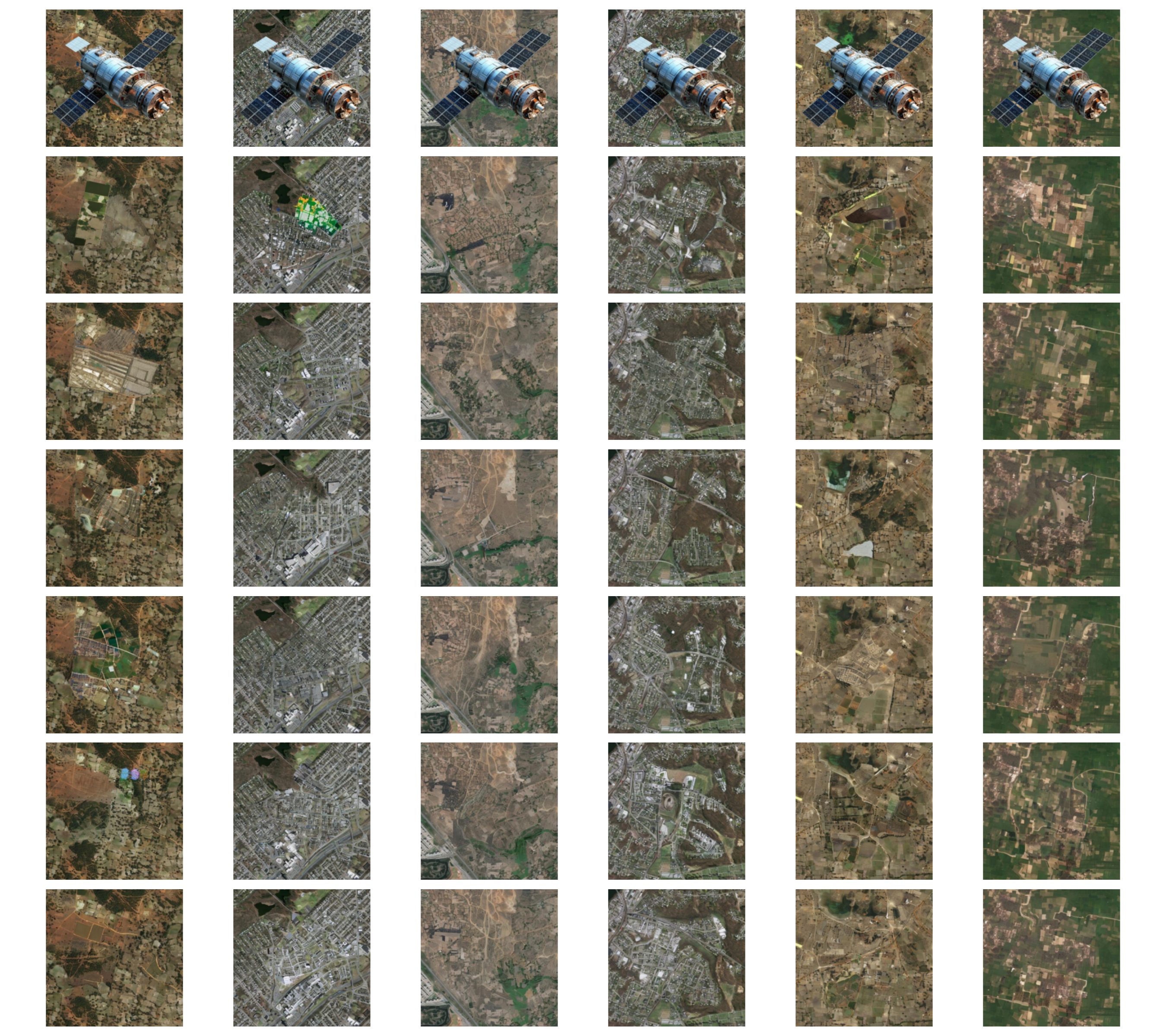}
    \caption{
    Side-by-side comparison of inpainting models applied to satellite imagery with complex agricultural occlusions. Each column represents a different region with structured land use and natural boundaries. The outputs (rows 2–7) from Stable Diffusion~\cite{rombach2022high}, RePaint~\cite{lugmayr2022repaint}, SatDiff~\cite{panboonyuen2025satdiff}, DPS~\cite{chung2022diffusion}, PSLD~\cite{rout2023solving}, and our method KAO are evaluated for realism and structural integrity. KAO distinctly preserves topographic layouts and subtle farming patterns—such as contour-aligned furrows and segmented plots—reflecting real-world agricultural logic. This substantiates KAO’s value for practical deployment in geospatial analysis and precision farming tasks.
    }
    \label{fig:inpainting_agriculture3}
\end{figure*}

To further evaluate the effectiveness of our proposed method \textbf{KAO} for satellite image inpainting, we present an in-depth visual analysis in agricultural contexts. Figures~\ref{fig:inpainting_agriculture1},~\ref{fig:inpainting_agriculture2}, and~\ref{fig:inpainting_agriculture3} (Appendix) showcase side-by-side comparisons across three geographically diverse samples dominated by structured farmland, cultivated plots, and irrigation patterns.

Each figure includes six occluded satellite scenes (columns), with outputs generated by leading diffusion-based and autoregressive baselines: Stable Diffusion~\cite{rombach2022high}, RePaint~\cite{lugmayr2022repaint}, SatDiff~\cite{panboonyuen2025satdiff}, DPS~\cite{chung2022diffusion}, PSLD~\cite{rout2023solving}, and our method KAO. Rows are structured consistently: the first row shows masked inputs, and the second through seventh rows show predictions from each method, respectively.

\paragraph{Reconstruction Fidelity:} 
Across all three figures, KAO exhibits superior capability in accurately reconstructing fine-grained agricultural structures. Unlike prior methods, which frequently generate blurred field boundaries or homogenized textures, KAO recovers highly structured patterns such as crop lines, plow contours, and segmented land plots. This is particularly evident in Figure~\ref{fig:inpainting_agriculture2}, where KAO correctly infers and reconstructs complex interleaved farmland layouts with minimal deviation from the ground truth.

\paragraph{Semantic Consistency:} 
KAO leverages its kernel-adaptive optimization and latent-space conditioning to better align inpainted content with surrounding visual semantics. For instance, in Figure~\ref{fig:inpainting_agriculture1}, KAO retains the original alignment and texture frequency of cultivated rows—features that are often distorted or lost in other methods like RePaint or Stable Diffusion. The ability to maintain semantic continuity is essential for downstream geospatial analytics, where accuracy of land cover classification is highly sensitive to spatial coherence.

\paragraph{Edge and Boundary Preservation:} 
A critical challenge in satellite inpainting lies in restoring sharp transitions between heterogeneous regions (e.g., crop-to-road or field-to-forest boundaries). KAO consistently produces sharper and more realistic borders compared to its counterparts. In Figure~\ref{fig:inpainting_agriculture3}, for example, KAO precisely reconstructs boundary alignments and water channel edges, which are either smoothed out or incorrectly extended in DPS and SatDiff outputs.

\paragraph{Generalization to Diverse Agricultural Typologies:} 
Each figure represents different types of agricultural settings: monoculture fields, mixed-use rural zones, and irrigation-based systems. KAO generalizes well across these typologies, adapting its inpainting strategy to preserve relevant features such as furrow direction, field segmentation, and vegetation density.

\paragraph{Why KAO Excels:} 
The presented qualitative results demonstrate that KAO is not only competitive with existing inpainting methods on general benchmarks but also particularly effective for agricultural satellite imagery—an application domain that requires high structural accuracy and strong contextual understanding. By integrating kernel-adaptive updates with Transformer-based latent refinement, KAO captures both fine-grained local textures and broad spatial dependencies, leading to superior reconstruction quality in terms of both fidelity and visual coherence.

Figures~\ref{fig:inpainting_agriculture1}--\ref{fig:inpainting_agriculture3} further illustrate KAO’s capability to reconstruct occluded agricultural features with high accuracy. This performance is essential for downstream tasks such as crop monitoring, land-use assessment, and environmental analysis, underscoring KAO’s potential as a robust solution for remote sensing in real-world scenarios.

\subsection{Novelty, Kernel Adaptation, Limitations, and TPT Motivation}

Regarding limitations, KAO is primarily optimized for VHR satellite imagery and shows reduced performance on medium- and low-resolution datasets (e.g., LANDSAT-8, 30m/pixel) due to the lack of fine-grained textures, which diminishes the effectiveness of hierarchical kernel weighting. Noise characteristics also differ, with atmospheric haze dominating over sensor-induced high-frequency noise. Mitigation strategies, such as multi-scale training and resolution-adaptive bandwidth parameters, are discussed to improve robustness across resolutions, supported by related works \cite{feng2024sa,ding2025slcgc}.

Finally, we clarify the motivation for employing the Token Pyramid Transformer (TPT) over alternative multi-scale architectures such as U-Net or HRNet. TPT preserves token-level granularity across scales without requiring downsampling/upsampling, unlike U-Net, and is computationally more efficient than HRNet while maintaining hierarchical semantic representations. Importantly, TPT naturally aligns with KAO's kernel-adaptive latent updates, as token-level features provide an ideal interface for adaptive kernel weighting \cite{zhang2022topformer}.

\end{document}